\documentclass[sn-basic,iicol]{sn-jnl}
\usepackage{graphicx}%
\usepackage{multirow}%
\usepackage{amsmath,amssymb,amsfonts}%
\usepackage{amsthm}%
\usepackage{mathrsfs}%
\usepackage[title]{appendix}%
\usepackage{xcolor}%
\usepackage{textcomp}%
\usepackage{manyfoot}%
\usepackage{booktabs}%
\usepackage{algorithm}%
\usepackage{algorithmicx}%
\usepackage{algpseudocode}%
\usepackage{listings}%
\usepackage{natbib}
\usepackage{url}
\usepackage{mathtools}
\usepackage{balance}
\usepackage{hhline}
\usepackage{makecell}
\usepackage{epsfig}
\usepackage{subcaption}
\usepackage{multirow}
\usepackage{makecell}
\usepackage{xcolor, soul}
\usepackage{array}
\usepackage{hhline}

\begin{document}

\title[Article Title]{One-shot Neural Face Reenactment via Finding Directions in GAN's Latent Space}


\author*[1]{\fnm{Stella} \sur{Bounareli}}\email{k2033759@kingston.ac.uk}

\author[2]{\fnm{Christos} \sur{Tzelepis}}\email{Christos.Tzelepis@city.ac.uk}

\author[1]{\fnm{Vasileios} \sur{Argyriou}}\email{vasileios.argyriou@kingston.ac.uk}

\author[3]{\fnm{Ioannis} \sur{Patras}}\email{i.patras@qmul.ac.uk}

\author[3]{\fnm{Georgios} \sur{Tzimiropoulos}}\email{g.tzimiropoulos@qmul.ac.uk}

\affil[1]{\orgdiv{School of Computer Science and Mathematics}, \orgname{Kingston University}, \orgaddress{\street{55-59 Penrhyn Road}, \city{London}, \postcode{KT1 2EE}, \country{United Kingdom}}}

\affil[2]{\orgdiv{School of Science and Technology}, \orgname{City University of London}, \orgaddress{\street{Northampton Square}, \city{London}, \postcode{EC1V 0HB}, \country{United Kingdom}}}

\affil[3]{\orgdiv{School of Electronic Engineering and Computer Science}, \orgname{Queen Mary University of London}, \orgaddress{\street{Mile End Road}, \city{London}, \postcode{E1 4NS}, \country{United Kingdom}}}

% %\authorrunning{Short form of author list} % if too long for running head

%%==================================%%
%%             abstract             %%
%%==================================%%

\abstract{In this paper, we present our framework for neural face/head reenactment whose goal is to transfer the 3D head orientation and expression of a target face to a source face. Previous methods focus on learning embedding networks for identity and head pose/expression disentanglement which proves to be a rather hard task, degrading the quality of the generated images. We take a different approach, bypassing the training of such networks, by using (fine-tuned) pre-trained GANs which have been shown capable of producing high-quality facial images. Because GANs are characterized by weak controllability, the core of our approach is a method to discover which directions in latent GAN space are responsible for controlling head pose and expression variations. We present a simple pipeline to learn such directions with the aid of a 3D shape model which, by construction, inherently captures disentangled directions for head pose, identity, and expression. Moreover, we show that by embedding real images in the GAN latent space, our method can be successfully used for the reenactment of real-world faces. Our method features several favorable properties including using a single source image (one-shot) and enabling cross-person reenactment. Extensive qualitative and quantitative results show that our approach typically produces reenacted faces of notably higher quality than those produced by state-of-the-art methods for the standard benchmarks of VoxCeleb1 \& 2.}




\maketitle

%%%%%%%%%%%%%%%%%%%%%%%%%%%%%%%%%%%%%%%%%%%%%%%%%%%%%%%%%%%%%%%%%%%%%%%%%%%%%%%%
%%                                                                            %%
%%                              [ Introduction ]                              %%
%%                                                                            %%
%%%%%%%%%%%%%%%%%%%%%%%%%%%%%%%%%%%%%%%%%%%%%%%%%%%%%%%%%%%%%%%%%%%%%%%%%%%%%%%%
\section{Introduction}\label{sec:introduction}

    Neural face reenactment aims to transfer the rigid 3D face/head orientation \textit{and} the deformable facial expression of a target facial image to a source facial image. Such technology is the key enabler for creating high-quality digital head avatars that find a multitude of applications in telepresence, Augmented Reality/Virtual Reality (AR/VR), and the creative industries. Recently, thanks to the advent of Deep Learning, Neural Face Reenactment has seen remarkable progress~\cite{burkov2020neural,zakharov2020fast,wang2021one,meshry2021learned}. In spite of this, synthesizing photorealistic face/head sequences remains a challenging problem with the quality of existing solutions being far from sufficient for the demanding aforementioned applications.

    A major challenge that most prior works~\cite{bao2018towards, zeng2020realistic, zakharov2019few,zakharov2020fast,burkov2020neural,ha2020marionette} have focused on is how to achieve identity and head pose/expression disentanglement to both preserve the appearance and identity characteristics of the source face and successfully transfer the head pose and the expression of the target face. A recent line of research relies on training conditional Generative Adversarial Networks (GANs)~\cite{deng2020disentangled,KowalskiECCV2020,shoshan2021gan} in order to produce disentangled embeddings and control the generation process. However, such methods mainly focus on synthetic image generation, rendering reenactment on real faces challenging. Another line of works~\cite{zakharov2019few, zakharov2020fast} rely on training with paired data (i.e., source and target facial images of the same identity), leading to poor cross-person face reenactment.

    In this work, we propose a neural face reenactment framework that addresses the aforementioned limitations of state-of-the-art (SOTA), motivated by the remarkable ability of modern pre-trained GANs (e.g., StyleGAN~\cite{karras2019style,karras2020analyzing,karras2020training}) in generating realistic and aesthetically pleasing faces, often indistinguishable from real ones. The research question we address in this paper is: \textit{Can a pre-trained GAN be adapted for face reenactment?} A key challenge that needs to be addressed to this end is the absence of any inherent semantic structure in the latent space of GANs. In order to gain control over the generative process, inspired by~\cite{voynov2020unsupervised}, we propose to learn a set of latent direction (i.e., direction vectors in the GAN's latent space) that are responsible for controlling head pose and expression variations in the generated facial images. Knowledge of these directions directly equips the pre-trained GAN with the ability of controllable generation in terms of head pose and expression, allowing for effective face reenactment. Specifically, in this work we present a simple pipeline to learn such directions leveraging the ability of a linear 3D shape model~\cite{feng2020deca} in capturing disentangled directions for head pose, identity, and expression, which is crucial towards effective neural face reenactment. Moreover, another key challenge that needs to be addressed is how to use the GAN for the manipulation of real-world images. Capitalizing on~\cite{tov2021designing}, we further show that by embedding real images in the GAN latent space, our pipeline can be successfully used for real face reenactment. Overall, we make the following contributions: 
    \begin{enumerate}
        \item Instead of training from-scratch conditional generative models~\cite{burkov2020neural,zakharov2020fast}, we present a novel approach to face reenactment by finding the directions in the latent space of a pre-trained GAN (i.e., StyleGAN2~\cite{karras2020analyzing} fine-tuned on the VoxCeleb1 dataset) that are responsible for controlling the rigid head orientation and expression, and show how these directions can be used for neural face reenactment on video datasets. 
        \item We present \textit{a simple pipeline} that is trained with the aid of a linear 3D shape model~\cite{feng2020deca}, that is inherently equipped with disentangled directions for facial shape in terms of head pose, identity and expression. We further show that our pipeline can be trained with real images by firstly embedding them into the GAN space, allowing for effective reenactment of real-world faces.
        \item We show that our method features several favorable properties including requiring a \textit{single source image} (one-shot), and enabling cross-person reenactment. 
        \item We perform several qualitative and quantitative comparisons with recent state-of-the-art reenactment methods, illustrating that our approach typically produces reenacted faces of notably higher quality for the standard benchmarks of VoxCeleb1 \& 2~\cite{Nagrani17,Chung18b}.
    \end{enumerate}

    Compared to our previous work in~\cite{bounareli2022finding}, this paper further investigates the real image inversion step and proposes a joint training scheme (Sect.~\ref{ssec:joint_training}) that eliminates the need for the optimization step during inference, described in Sect.~\ref{ssec:real}, resulting in a more efficient inference process and better quantitative and qualitative results. The proposed joint training scheme efficiently addresses existing visual artifacts on the reenacted images caused by large head pose variations between the source and target faces, resulting in improved overall image quality. We qualitatively and quantitatively show that by jointly learning the real image inversion encoder and the directions, our method achieves compelling results without the need of one-shot fine-tuning during inference. Finally, to further improve the visual quality of the reenacted images in terms of crucial (for the purpose of face reenactment) background and identity characteristics, we propose to further fine-tune the feature space $\mathcal{F}$ of StyleGAN2 (Sect.~\ref{ssec:feature_space}).

%%%%%%%%%%%%%%%%%%%%%%%%%%%%%%%%%%%%%%%%%%%%%%%%%%%%%%%%%%%%%%%%%%%%%%%%%%%%%%%%
%%                                                                            %%
%%                              [ Related work ]                              %%
%%                                                                            %%
%%%%%%%%%%%%%%%%%%%%%%%%%%%%%%%%%%%%%%%%%%%%%%%%%%%%%%%%%%%%%%%%%%%%%%%%%%%%%%%%
\section{Related work}\label{sec:related_work}

    \subsection{Semantic face editing}
        Several recent works~\cite{shen2020interfacegan,2020ganspace,voynov2020unsupervised,shen2020closedform, oldfield2021tensor,tzelepis2021warpedganspace,yao2021latent,yang2021discovering, oldfield2022panda,tzelepis2022contraclip} study the existence of directions/paths in the latent space of a pre-trained GAN in order to perform editing (i.e., with respect to specific facial attributes) on the generated facial images. \cite{voynov2020unsupervised} introduced an unsupervised method that optimizes a set of vectors in the GAN's latent space by learning to distinguish (using a ``reconstructor'' network) the image transformations caused by distinct latent directions. This leads to the discovery of a set of ``interpretable'', but not ``controllable'', directions -- i.e., the optimized latent directions cannot be used for controllable (in terms of head pose and expression) facial editing and, thus, for face reenactment. Our method is inspired by the work of~\cite{voynov2020unsupervised}, extending it in several ways to make it suitable for neural face reenactment. Another line of recent works allows for explicit controllable facial image editing ~\cite{deng2020disentangled,gif, durall2021facialgan, shoshan2021gan, wang2021cross, nitzan2020face, abdal2021styleflow}. However, these methods mostly rely on synthetic image editing rather than performing face reenactment on real video data. A work that is related to our framework is StyleRig~\cite{tewari2020stylerig}, which uses 3D Morphable Model's (3DMM)~\cite{blanz1999morphable} parameters to control the generated images from a pre-trained StyleGAN2~\cite{karras2020analyzing}. However, by contrast to our method, StyleRig's training pipeline is not end-to-end and is significantly more complicated than ours, while in order to learn better disentangled directions, StyleRig requires the training of distinct models for different attributes (e.g., head pose and expression). This, along with the fact that StyleRig operates mainly on synthetic images, poses a notable restriction towards real-world face reenactment, where various facial attributes change simultaneously. By contrast, we propose to learn all disentangled directions for face reenactment simultaneously, allowing in this way for the effective editing of all, a subset, or a single attribute, whilst we optimize our framework on real faces as well. A follow-up work, PIE~\cite{tewari2020pie}, focuses on inverting real images to enable editing using StyleRig~\cite{tewari2020stylerig}. However, their method is computationally expensive (10 min/image) which is prohibitive for video-based facial reenactment. By contrast, we propose a framework that effectively and efficiently performs face reenactment (0.13 sec/image).

    \subsection{GAN inversion}
        GAN inversion methods aim to encode real images into the latent space of pre-trained GANs~\cite{karras2019style,karras2020analyzing}, allowing for subsequent editing using existing methods of synthetic image manipulation. The major challenge in the GAN inversion problem comprises of the so called ``editability-perception'' trade-off; that is, finding a sweet spot between faithful reconstruction of the real image and the editability of the corresponding latent code. The majority of recent inversion methods~\cite{alaluf2021restyle,richardson2021encoding,tov2021designing, alaluf2022hyperstyle, dinh2022hyperinverter,wang2022high} train encoder-based architectures that focus on predicting the latent codes $\mathbf{w}$ that best reconstruct the original (real) images and that allow for subsequent editing. \cite{zhu2020domain} propose a hybrid approach which consists of learning an encoder followed by an optimization step on the latent space to refine the similarity between the reconstructed and real images. \cite{richardson2021encoding} introduce a method that aims to improve the ``editability-perception'' trade-off, while recently~\cite{roich2021pivotal} propose to fine-tune the generator to better capture/transfer appearance features.

        The aforementioned works typically perform inversion onto the $\mathcal{W}+$ latent space of StyleGAN2. However, \cite{parmar2022spatially} have shown that $\mathcal{W}+$ is not capable of fully reconstructing the real images. Specifically, details such as the background, the hair style or facial accessories i.e., hats and glasses, cannot be inverted with high fidelity. A recent line of works~\cite{wang2022high,xuyao2022,bai2022high,alaluf2022hyperstyle} propose to mitigate this by investigating more expressive spaces of StyleGAN2 (such as the feature space $\mathcal{F} \in\mathbb{R}^{h \times w \times c}$~\cite{kang2021gan}) to perform real image inversion. Although such methods are able to produce high quality reconstructions, their ability to accurately edit the inverted images is limited. Especially when changing the head pose, such methods tend to produce many visual artifacts (Fig.~\ref{fig:comparisons_inversion}). In order to balance between expressive invertibility and editing performance, the authors of~\cite{parmar2022spatially} (SAM) propose to fuse different spaces, i.e., the $\mathcal{W}+$ latent space and the feature space $\mathcal{F} = \{\mathcal{F}_4, \mathcal{F}_6, \mathcal{F}_8, \mathcal{F}_{10} \}$, where each one corresponds to a different feature layer of StyleGAN2~\cite{karras2020analyzing}. In more detail, they propose to break the facial images into different segments (background, hat, glasses etc.) and choose the most suitable space to invert each segment, leveraging the editing capabilities of the $\mathcal{W}+$ latent space and the reconstruction quality of the feature space $\mathcal{F}$. However, when performing global editings, i.e., changing the head pose orientation, SAM~\cite{parmar2022spatially} results in notable visual artifacts, in contrast to our method, as will be shown in the experimental section.

        \begin{figure*}[t]
            \begin{center}
            \includegraphics[width=1.0\textwidth]{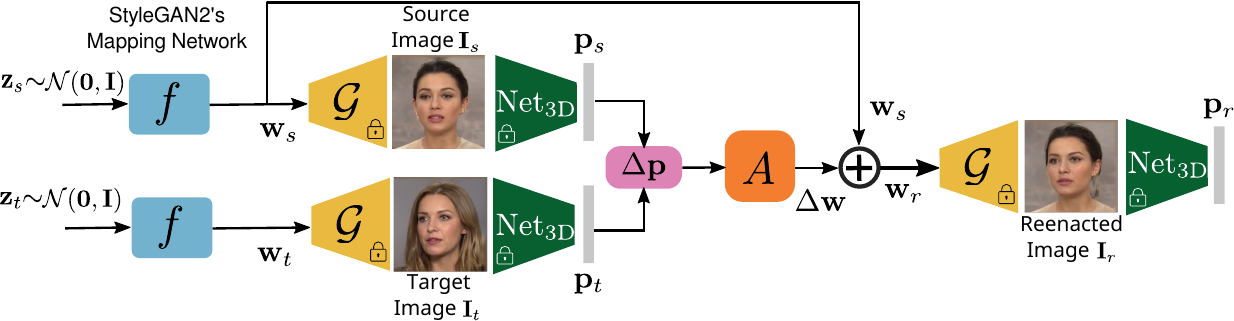}
            \end{center}
            \caption{\textbf{Overview of the proposed framework:} Given a pair of source $\mathbf{I}_s$ and target $\mathbf{I}_t$ images, we calculate the head pose/expression parameter vectors $\mathbf{p}_s$ and $\mathbf{p}_t$ using the $\mathrm{Net_{3D}}$ network, respectively. The matrix of directions $\mathbf{A}$ is trained so as, given the shift $\Delta \mathbf{w} = \mathbf{A}\Delta \mathbf{p}$, the reenacted image $\mathbf{I}_r$ generated using the latent code $\mathbf{w}_r = \mathbf{w}_s + \boldsymbol{\Delta}\mathbf{w}$, transfers the head pose and the expression of the target face, maintaining at the same time the identity of the source face.}
            \label{fig:architecture}
        \end{figure*}

    \subsection{Neural face reenactment}
      
        Neural face reenactment poses a challenging problem that requires strong generalization ability across many different identities and a large range of head poses and expressions. Many of the proposed methods rely on facial landmark information \cite{zakharov2019few, tripathy2020icface, zhang2020freenet,ha2020marionette, tripathy2021facegan, zakharov2020fast, wang2022latent, hsu2022dual}. Specifically, \cite{zakharov2020fast} propose an one-shot face reenactment method driven by landmarks, which decomposes an image on pose-dependent and pose-independent components. A limitation of landmark based methods is that landmarks preserve identity information, thus impeding their applicability on cross-subject face reenactment~\cite{burkov2020neural}. In order to mitigate this limitation~\cite{hsu2022dual} propose to use an ID-preserving Shape Generator (IDSG) that transforms the target facial landmarks so that they preserve the identity, i.e. facial shape, of the source image. Additionally, several methods~\cite{doukas2020headgan,yao2020mesh,ren2021pirenderer,yang2022face2face} rely on 3D shape models to remove the identity details of the driving images. Warping-based methods~\cite{wiles2018x2face,siarohin2019first,wang2021one,ren2021pirenderer, doukas2020headgan,yang2022face2face} synthesize the reenacted images based on the motion of the driving faces. Specifically, HeadGAN~\cite{doukas2020headgan} and Face2Face~\cite{yang2022face2face} are warping-based methods conditioned on 3D Morphable Models. Whilst such methods produce realistic results, they suffer from several visual artifacts and head pose mismatch, especially in large head pose variations. Finally,~\cite{meshry2021learned} propose a two-step architecture that aims to disentangle the spatial and style components of an image that leads to better preservation of the source identity.

        By contrast to the methods discussed above, which rely on the \textit{training of conditional generative models} on large \textit{paired} datasets in order to learn facial descriptors with disentanglement properties, in this paper we propose a novel and simple face reenactment framework that learns disentangled directions in the latent space of a StyleGAN2~\cite{karras2020analyzing} pre-trained on the VoxCeleb~\cite{Nagrani17} dataset. We show that the discovery of meaningful and disentangled directions that are responsible for controlling the head pose and the facial expression can be used for high quality self- and cross-identity reenactment.

%%%%%%%%%%%%%%%%%%%%%%%%%%%%%%%%%%%%%%%%%%%%%%%%%%%%%%%%%%%%%%%%%%%%%%%%%%%%%%%%
%%                                                                            %%
%%                            [ Proposed Method ]                             %%
%%                                                                            %%
%%%%%%%%%%%%%%%%%%%%%%%%%%%%%%%%%%%%%%%%%%%%%%%%%%%%%%%%%%%%%%%%%%%%%%%%%%%%%%%%
\section{Proposed Method}\label{sec:proposed_method}

    In this section, we present the proposed framework for one-shot neural face reenactment via finding directions in the latent space of StyleGAN2. More specifically, we begin with the most basic variant of our framework for finding reenactment latent directions using unpaired synthetic images in Sect.~\ref{ssec:latent} -- an overview of this is shown in Fig.~\ref{fig:architecture}. Next, in Sect.~\ref{ssec:real} we extend this methodology for handling real images along with synthetic ones (i.e., towards real face reenactment), while in Sect.~\ref{ssec:video} we investigate the incorporation of paired video data. In Sect.~\ref{ssec:joint_training} we introduce a joint training scheme that allows for optimization-free reenactment, leading to efficient and consistent neural reenactment. Finally, in Sect.~\ref{ssec:feature_space}, on top of the previously introduced variants of our method, we propose the refinement of crucial visual details (i.e., background, hair style) by leveraging the impressive reconstruction capability of StyleGAN2's feature space $\mathcal{F}$.

    %%%%%%%%%%%%%%%%%%%%%%%%%%%%%%%%%%%%%%%%%%%%%%%%%%%%%%%%%%%%%%%%%%%%%%%%%%%%%%%%
    %%                [ Finding the reenactment latent directions ]               %%
    %%%%%%%%%%%%%%%%%%%%%%%%%%%%%%%%%%%%%%%%%%%%%%%%%%%%%%%%%%%%%%%%%%%%%%%%%%%%%%%%
    \subsection{Finding reenactment latent directions on unpaired synthetic images} \label{ssec:latent}
    
        % +--------------------------------------------------------------------------+ %
        % |                         StyleGAN2 background                             | %
        % +--------------------------------------------------------------------------+ %
        \subsubsection{StyleGAN2 background}\label{sssec:stylegan2}

            Let $\mathcal{G}$ denote the generator of StyleGAN2~\cite{karras2020analyzing}, as shown in Fig.~\ref{fig:architecture}. Specifically, $\mathcal{G}$ takes as input a latent code $\mathbf{w}\in\mathcal{W}\subset\mathbb{R}^{512}$, which is typically the output of StyleGAN2's input MLP-based Mapping Network $f$ that acts on samples $\mathbf{z}\in\mathbb{R}^{512}$ drawn from the standard Gaussian $\mathcal{N}(\mathbf{0},\mathbf{I})$. That is, given a latent code $\mathbf{z}\sim\mathcal{N}(\mathbf{0},\mathbf{I})$, the generator produces an image $\mathcal{G}(f(\mathbf{z}))\in\mathbb{R}^{3\times256\times256}$.

            StyleGAN2 is typically pre-trained on the Flickr-Faces-HQ (FFHQ) dataset~\cite{karras2019style}, which exhibits poor diversity in terms of head pose and facial expression; for instance, FFHQ does not typically account for roll changes in head pose. In order to compare our method with other state-of-the-art methods, we fine-tune StyleGAN2's generator $\mathcal{G}$ on the VoxCeleb dataset~\cite{Nagrani17}, which provides a much wider range of head poses and facial expressions, rendering it very useful for the task of neural face reenactment by finding the appropriate latent directions as will be discussed in the following sections. We note that we fine-tune the StyleGAN2's generator on VoxCeleb dataset using the method provided by~\cite{karras2020training}, while we do not impose any reenactment objectives. That is, the fine-tuned generator can produce synthetic images with random identities (different from the identities of VoxCeleb) that follow the distribution of VoxCeleb dataset in terms of head poses and expressions. 

        % +--------------------------------------------------------------------------+ %
        % |                             3DMM (Net3D)                                 | %
        % +--------------------------------------------------------------------------+ %
        \subsubsection{3D Morphable Model (Net3D)}\label{sssec:net3d}
            
            Given an image, Net3D~\cite{feng2020deca} encodes the depicted face's pose into a facial shape vector $\textbf{s}\in\mathbb{R}^{3N}$, where $N$ denotes the number of vertices, which can be decomposed in terms of a linear 3D facial shape as
            \begin{equation}\label{eq:3dmm}
                \textbf{s} = \bar{\textbf{s}} + \textbf{S}_{i}\mathbf{p}_{i} +  \textbf{S}_{\theta}\mathbf{p}_{\theta} + \textbf{S}_{e}\mathbf{p}_{e},
            \end{equation}
            where $\bar{\textbf{s}}$ denotes the mean 3D facial shape, $\textbf{S}_{i}\in\mathbb{R}^{3N\times m_{i}}$, $\textbf{S}_{\theta}\in\mathbb{R}^{3N\times m_{\theta}}$ and $\textbf{S}_{e}\in\mathbb{R}^{3N\times m_{e}}$ denote the PCA bases for identity, head orientation and expression, and $\mathbf{p}_{i}$, $\mathbf{p}_{\theta}$ and $\mathbf{p}_{e}$ denote the corresponding identity, head orientation and expression coefficients, respectively. The variables $m_{i}$, $m_{\theta}$ and $m_{e}$ correspond to the number of identity, head pose and expression coefficients. For reenactment, we are interested in manipulating head orientation and expression, thus, our head pose/expression parameter vector is given as $\mathbf{p}=[\mathbf{p}_{\theta}, \mathbf{p}_{e}] \in \mathbb{R}^{3+m_e}$. We note that all PCA shape bases are orthogonal to each other, and hence they capture disentangled variations of identity and expression. Finally, we note that they are calculated in a frontalized reference frame, thus, they are also disentangled from head orientation. These bases can be also interpreted as directions in the shape space. We propose to learn similar directions in the GAN latent space as described in detail in the following section. 
            
        % +--------------------------------------------------------------------------+ %
        % |                           Reenactment latent directions                  | %
        % +--------------------------------------------------------------------------+ %
        \subsubsection{Reenactment latent directions}\label{sssec:latent_directions}
            
            In particular, we propose to associate a change $\Delta \mathbf{p}$ in the head pose orientation and expression, with a change $\Delta \mathbf{w}$ in the (intermediate) latent GAN space so that the two generated images $G(\mathbf{w})$ and $G(\mathbf{w}+\Delta \mathbf{w})$ differ only in head pose and expression by the same amount $\Delta \mathbf{s}$ induced by $\Delta \mathbf{p}$. If the directions sought in the GAN latent space are assumed to be linear~\cite{nitzan2021large}, this implies the following linear relationship
            \begin{equation}\label{Eq:Dw}
                \Delta \mathbf{w} = \mathbf{A}\Delta \mathbf{p},
            \end{equation}
            where $\mathbf{A}\in \mathbb{R}^{d_{\mathrm{out}}\times d_{\mathrm{in}}}$ is a matrix, the columns of which represent the directions in GAN latent space. In our case, $d_{\mathrm{in}} = (3+m_e)$ and $d_{\mathrm{out}} = N_{l}\times512$, where $N_{l}$ is the number of the generator's layers we opt to apply shift changes.

        % +--------------------------------------------------------------------------+ %
        % |                               Training                                   | %
        % +--------------------------------------------------------------------------+ %
        \subsubsection{Training pipeline}\label{sssec:training}

            In order to optimize the matrix of controllable latent directions $\mathbf{A}$, we propose a simple pipeline, shown in Fig.~\ref{fig:architecture}. Specifically, during training, a pair of a source ($\mathbf{z}_s$) and a target ($\mathbf{z}_t$) latent codes are drawn from $\mathcal{N}(\mathbf{0},\mathbf{I})$, giving rise to a pair of a source ($\mathbf{I}_s = \mathcal{G}(f(\mathbf{z}_s))$) and a target ($\mathbf{I}_t = \mathcal{G}(f(\mathbf{z}_t))$) images, as shown in the left part of Fig.~\ref{fig:architecture}. The pair of images $\mathbf{I}_s$ and $\mathbf{I}_t$ are then encoded by the pre-trained Net3D into the head pose/expression parameter vectors $\mathbf{p}_s$ and $\mathbf{p}_t$, respectively. Using (\ref{Eq:Dw}), we calculate the shift $\Delta\mathbf{w}$ in the intermediate latent space of StyleGAN2 as $\Delta\mathbf{w}=\mathbf{A}\Delta\mathbf{p}=\mathbf{A}(\mathbf{p}_t-\mathbf{p}_s)$ and the \textit{reenactment latent code} $\mathbf{w}_{r}=\mathbf{w}_s+\Delta\mathbf{w}$. Using the latter we arrive at the reenacted image $\mathbf{I}_{r}= \mathcal{G}(\mathbf{w}_{r})$. 
           
            It is worth noting that the only trainable module of the proposed framework is the matrix $\mathbf{A}\in \mathbb{R}^{d_{\mathrm{out}}\times d_{\mathrm{in}}}$ -- i.e., the number of trainable parameters of the proposed framework is $65K$. We also note that, before training, we estimate the distribution of each element of the head pose/expression parameters $\mathbf{p}$ by randomly generating $10\mathrm{K}$ images and calculating using the pre-trained Net3D~\cite{feng2020deca} their corresponding $\mathbf{p}$ vectors. Using the estimated distributions, during training, we re-scale each element of $\mathbf{p}$ from its original range to a common range $[-a,a]$ ($a$ being a hyperparameter empirically set to $6$). In the appendices (Sect.~\ref{subsubsec:facial_pose}) we further discuss the re-scaling of each element of $\mathbf{p}$. To further encourage disentanglement in the optimized latent directions matrix $\mathbf{A}$, we follow a training strategy where for $50\%$ of the training samples we reenact only one attribute by using $\Delta \mathbf{p} = [0,\ldots,\varepsilon,\ldots,0]$, where $\varepsilon$ is uniformly sampled from $\mathcal{U}[-a, a]$. In the appendices (Sect.~\ref{subsubsec:disent}) we show that the above training strategy improves the disentanglement between the learned directions.

        % +--------------------------------------------------------------------------+ %
        % |                                 Losses                                   | %
        % +--------------------------------------------------------------------------+ %
        \subsubsection{Losses}\label{sssec:losses}

            We train our framework by minimizing the following total loss:
            \begin{equation}\label{eq:loss_all}
                \mathcal{L} = \lambda_{r} \mathcal{L}_{r} + \lambda_{id} \mathcal{L}_{id} + \lambda_{per} \mathcal{L}_{per},
            \end{equation}
            where $\mathcal{L}_{r}$, $\mathcal{L}_{id}$, and $\mathcal{L}_{per}$ denote respectively the \textit{reenactment}, \textit{identity}, and \textit{perceptual} losses with $\lambda_{r}$, $\lambda_{id}$, and $\lambda_{per}$ being weighting hyperparameters empirically set to $\lambda_{r}=1$,  $\lambda_{id}=10$ and $\lambda_{per}=10$. We detail each loss term below.

            \paragraph{Reenactment loss $\mathcal{L}_{r}$} We define the reenactment loss as
            % $$
            %     \mathcal{L}_{r} = \mathcal{L}_{sh} + \mathcal{L}_{eye} + \mathcal{L}_{mouth},
            % $$
            \begin{equation}\label{eq:reenactment_loss}
               \mathcal{L}_{r} = \mathcal{L}_{sh} + \mathcal{L}_{eye} + \mathcal{L}_{mouth},
            \end{equation}
            where the \textit{shape} loss term $\mathcal{L}_{sh}=\|\mathbf{S}_r-\mathbf{S}_{gt}\|_1$ imposes head pose and expression transfer from target to source, where $\mathbf{S}_r$ is the 3D shape of the reenacted image and $\mathbf{S}_{gt}$ is the reconstructed \textit{ground-truth} 3D shape calculated using (\ref{eq:3dmm}). Specifically, the ground-truth 3D facial shape $\mathbf{S}_{gt}$ should have the identity, i.e., facial shape, of the source image and the facial expression and head pose of the target image, either on the task of self reenactment or on cross-subject reenactment. On self reenactment $\mathbf{S}_{gt}$ is the same with $\mathbf{S}_{t}$, where $\mathbf{S}_{t}$ is the facial shape of the target image. On cross-subject reenactment, we calculate $\mathbf{S}_{gt}$  using the identity coefficients $\mathbf{p}_{i}^s$ of the source face and the facial expression and head pose coefficients $\mathbf{p}_{e}^t, \mathbf{p}_{\theta}^t$ of the target face as:
             \begin{equation}\label{eq:3dmm_ground}
                \textbf{S}_{gt} = \bar{\textbf{S}} + \textbf{S}_{i}\mathbf{p}_{i}^s +  \textbf{S}_{\theta}\mathbf{p}_{\theta}^t + \textbf{S}_{e}\mathbf{p}_{e}^t,
            \end{equation}
            To enhance the expression transfer, we calculate the \textit{eye} ($\mathcal{L}_{eye}$) and the \textit{mouth} ($\mathcal{L}_{mouth}$) losses. The eye loss $\mathcal{L}_{eye}$ (the mouth loss $\mathcal{L}_{mouth}$ is computed in a similar fashion) compares the inner distances between the eye landmark pairs of upper and lower eyelids between the reenacted and reconstructed ground-truth shapes. In Appendix~\ref{subsec:shape_losses}, we provide a detailed discussion on $\mathcal{L}_{eye}$ and $\mathcal{L}_{mouth}$.

            \paragraph{Identity loss $\mathcal{L}_{id}$} We define the identity loss as the cosine similarity between feature representations extracted from the source $\mathbf{I}_s$ and the reenacted $\mathbf{I}_r$ image using ArcFace~~\cite{deng2019arcface}. The identity loss imposes the identity preservation between the source and the reenacted image.

            \paragraph{Perceptual loss $\mathcal{L}_{per}$} We defined the perceptual loss similarly to~\cite{johnson2016perceptual} in order to improve the quality of the reenactment face images.

    %%%%%%%%%%%%%%%%%%%%%%%%%%%%%%%%%%%%%%%%%%%%%%%%%%%%%%%%%%%%%%%%%%%%%%%%%%%%%%%%
    %%                       [ Fine-tuning on real images ]                       %%
    %%%%%%%%%%%%%%%%%%%%%%%%%%%%%%%%%%%%%%%%%%%%%%%%%%%%%%%%%%%%%%%%%%%%%%%%%%%%%%%%
    \subsection{Fine-tuning on unpaired real images}\label{ssec:real}

        In this section, we extend the basic pipeline of the proposed framework, described in the previous section, in order to learn from both synthetic and real images. For doing so, we propose to (a) use a pipeline for inverting the images back to the latent code space of StyleGAN2, and (b) adopt a mixed training approach (using both synthetic and inverted latent codes) for discovering the latent directions (Sect.~\ref{sssec:latent_directions}).

        As discussed in previous sections, the main challenge in the GAN inversion problem is finding a good trade-off between faithful reconstruction of the real image and effective editability using the inverted latent code. Although satisfying both requirements is challenging~\cite{alaluf2021restyle,richardson2021encoding,tov2021designing}, we found that the following pipeline produces compelling results for the purposes of our goal (i.e., face/head reenactment). During training, we employ an encoder based method (e4e)~\cite{tov2021designing} to invert the real images into the $\mathcal{W}+$ latent space of StyleGAN2~\cite{abdal2019image2stylegan}. However, directly using the inverted $\mathcal{W}+$ latent codes performs poorly in face reenactment due to the domain gap between the synthetic and inverted latent codes. To alleviate this, we propose a mixed-data approach (i.e., using both synthetic and real images) for training the pipeline presented in Sect.~\ref{ssec:latent}. Specifically, we first invert the extracted frames from the VoxCeleb dataset, and during training, at each iteration (i.e., for each batch) we use 50\% random latent codes $\mathbf{w}$ and 50\% embedded latent codes $\mathbf{w}^{inv}$.

\begin{figure}[t]
        \begin{center}
        {\includegraphics[width=1.0\linewidth]{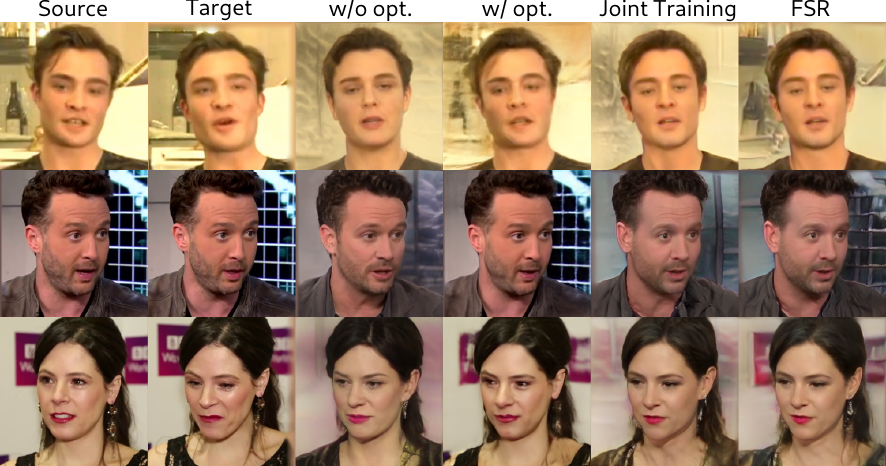}}
        \end{center}
          \caption{Examples of face reenactment without (``w/o opt.'') and with (``w/ opt.'') the generator's optimization. We additionally show results using our proposed joint training scheme (``Joint Training'') and the refinement of StyleGAN2's feature space (``FSR'') described in Sect.~\ref{ssec:joint_training} and~\ref{ssec:feature_space}, respectively.}
        \label{fig:inversion}
        \end{figure}

        Since the inverted images using e4e~\cite{tov2021designing} might still be missing some crucial identity details, we propose to use an additional optimization step (\textit{only during inference}), similarly to~\cite{roich2021pivotal}, in order to slightly update the generator $\mathcal{G}$ and arrive at better reenacted images in terms of identity preservation. Note that this step does not affect the calculation of $\mathbf{w}^{inv}$ and is used only during inference to obtain a higher quality inversion. We perform the optimization for 200 steps and only on the source frame of each video. In Fig.~\ref{fig:inversion} we illustrate examples of neural face reenactment without optimizing the generator's weights (w/o opt. -- third column) and with optimization (w/ opt. -- fourth column), where we observe that, clearly, the reenacted images without the additional optimization step are not able to faithfully reconstruct the real images, while the reenacted images after optimizing the generator weights resembles the real ones more closely.

    %%%%%%%%%%%%%%%%%%%%%%%%%%%%%%%%%%%%%%%%%%%%%%%%%%%%%%%%%%%%%%%%%%%%%%%%%%%%%%%%
    %%             [ Fine-tuning on paired real images (video data) ]             %%
    %%%%%%%%%%%%%%%%%%%%%%%%%%%%%%%%%%%%%%%%%%%%%%%%%%%%%%%%%%%%%%%%%%%%%%%%%%%%%%%%
    \subsection{Fine-tuning on paired real images (video data)}\label{ssec:video}

        In the previous sections, we presented the proposed framework for learning from unpaired synthetic and real images. Whilst this provides the benefit of learning from a very large number of identities, making it useful for cross-person reenactment, we show that we can achieve additional improvements by optimizing novel losses introduced by further training on paired data from the VoxCeleb1~\cite{Nagrani17} video dataset.
        
        Compared to training from scratch on video data, as in most previous methods (e.g.~\cite{zakharov2019few,zakharov2020fast,burkov2020neural}), we argue that our approach offers a more balanced strategy that combines the best of both worlds; that is, training with unpaired images and fine-tuning with paired video data. From each video of our training set, we randomly sample a source and a target face that have the same identity but different head pose/expression. Consequently, we minimize the following loss function 
        \begin{equation}\label{eq:loss_paired}
            \mathcal{L} = \lambda_{r}\mathcal{L}_{r} + \lambda_{id}\mathcal{L}_{id} + \lambda_{per}\mathcal{L}_{per} + \lambda_{pix}\mathcal{L}_{pix},
        \end{equation}
        where $\mathcal{L}_{r}$ is the same reenactment loss defined in Sect.~\ref{ssec:latent}, $\mathcal{L}_{id}$ and $\mathcal{L}_{per}$ are the identity and perceptual losses, however this time calculated between the reenacted $\mathbf{I}_r$ and the target image $\mathbf{I}_t$, and $\mathcal{L}_{pix}$ is a pixel-wise $L1$ loss between the reenacted and target images.

    %%%%%%%%%%%%%%%%%%%%%%%%%%%%%%%%%%%%%%%%%%%%%%%%%%%%%%%%%%%%%%%%%%%%%%%%%%%%%%%%
    %%                 [ Joint Training ]                     %%
    %%%%%%%%%%%%%%%%%%%%%%%%%%%%%%%%%%%%%%%%%%%%%%%%%%%%%%%%%%%%%%%%%%%%%%%%%%%%%%%%
    \subsection{Joint Training of the real image inversion encoder $\mathcal{E}_w$ and the directions matrix $\mathbf{A}$}\label{ssec:joint_training}

        \begin{figure}[t]
        \begin{center}
        \includegraphics[width=1.0\linewidth]{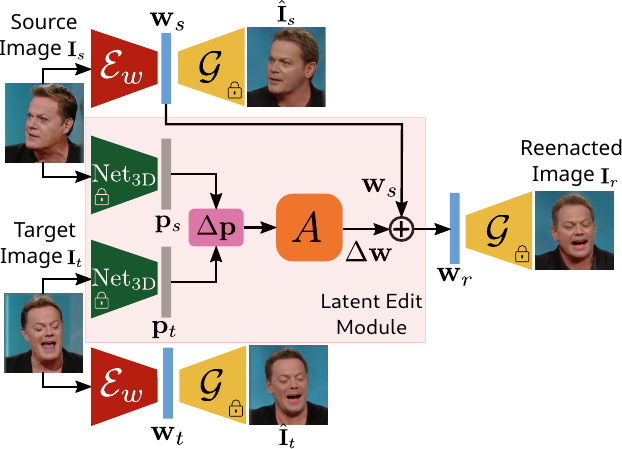}
        \end{center}
        \caption{To eliminate the need for the optimization step during inference, we propose to jointly train the real image inversion encoder $\mathcal{E}_w$ and the directions matrix $\mathbf{A}$. We note that during training both the generator $\mathcal{G}$ and the $\mathrm{Net_{3D}}$ network are frozen.} 
        \label{fig:architecture_joint}
        \end{figure}

        As discussed in Sect.~\ref{ssec:real}, the encoder-based e4e~\cite{tov2021designing} inversion method often fails to faithfully reconstruct real images by typically failing to preserve crucial identity characteristics, as shown in the third column (``w/o opt.'') of Fig.~\ref{fig:inversion}. Clearly, this poses a certain limitation to the face reenactment methodology presented in Sect.~\ref{sssec:training}. Optimizing the generator's weights leads to notable improvements (Sect.~\ref{ssec:real}), as shown in the fourth column (``w/ opt.'') of Fig.~\ref{fig:inversion}, albeit, this comes at a significant cost for the task of face reenactment (that is, the optimization of $\mathcal{G}$ takes approximately 20 sec. per frame). 
         
        In this section, we propose to \textit{jointly} train the real image inversion encoder $\mathcal{E}_w$ and the directions matrix $\mathbf{A}$, which leads to optimization-free face reenactment at inference time. For doing so, we use paired data as described in Sect.~\ref{ssec:video}. An overview of this approach is shown in Fig.~\ref{fig:architecture_joint}. Specifically, we first sample a source ($\mathbf{I}_s$) and a target ($\mathbf{I}_t$) image from the same video of VoxCeleb1~\cite{Nagrani17} training set, that have the same identity but different head pose/expression. Those images are then fed into the inversion encoder $\mathcal{E}_w$ to predict the corresponding source ($\mathbf{w}_s$) and target ($\mathbf{w}_t$) latent codes. Then, the pre-trained $\mathrm{Net_{3D}}$ network extracts the corresponding source ($\mathbf{p}_s$) and target ($\mathbf{p}_t$) parameter vectors. Finally, as described in Sect.~\ref{ssec:latent}, we generate the reenacted image using the latent code $\mathbf{w}_{r} = \mathbf{w}_s + \Delta \mathbf{w}$, where $\Delta \mathbf{w}=\mathbf{A}(\mathbf{p}_t-\mathbf{p}_s)$.

        % === Encoder optimization ===
        \subsubsection{Real image encoder $\mathcal{E}_w$ optimization objective}
        
            In order to train the real image encoder $\mathcal{E}_w$ we minimize the following loss:
            \begin{equation}\label{eq:recon_losses}
            \begin{split}
                \mathcal{L}_{\mathcal{E}_w} = \lambda_{id} (\mathcal{L}_{id}(\mathbf{I}_s, \hat{\mathbf{I}}_s) + \mathcal{L}_{id}(\mathbf{I}_t, \hat{\mathbf{I}}_t)) + \\
                \lambda_{per} (\mathcal{L}_{per}(\mathbf{I}_s, \hat{\mathbf{I}}_s) + \mathcal{L}_{per}(\mathbf{I}_t, \hat{\mathbf{I}}_t)) + \\
                 \lambda_{pix} (\mathcal{L}_{pix}(\mathbf{I}_s, \hat{\mathbf{I}}_s) + \mathcal{L}_{pix}(\mathbf{I}_t, \hat{\mathbf{I}}_t)) + \\
                  \lambda_{style} (\mathcal{L}_{style}(\mathbf{I}_s, \hat{\mathbf{I}}_s) + \mathcal{L}_{style}(\mathbf{I}_t, \hat{\mathbf{I}}_t)),
            \end{split}
            \end{equation}
            where $\mathcal{L}_{id}$, $\mathcal{L}_{per}$, and $\mathcal{L}_{pix}$ denote the identity, perceptual, and pixel-wise losses described in the previous sections.
            
            Additionally, to further improve the style and the quality of the reconstructed images we propose to use a style loss $\mathcal{L}_{style}$ similarly to~\cite{barattin2023attribute}. Specifically, we use FaRL~\cite{zheng2022general}, a method for general facial representation learning that leverages contrastive learning between images and text pairs to learn meaningful feature representations of facial images. In our method, we use the image Transformer-based encoder, $\mathcal{E}_{FaRL}$, to extract a 512-dimensional feature vector from each image. The proposed style loss is defined as:
            \begin{equation}
            \begin{split}
                \mathcal{L}_{style} = \| \mathcal{E}_{FaRL}(\mathbf{I}_s) - \mathcal{E}_{FaRL}(\hat{\mathbf{I}}_s)\|_1 + \\
                \| \mathcal{E}_{FaRL}(\mathbf{I}_t) - \mathcal{E}_{FaRL}(\hat{\mathbf{I}}_t)\|_1.
            \end{split}
            \end{equation}

        % ==== Matrix A optimization ===
        \subsubsection{Directions matrix $\mathbf{A}$ optimization objective}
        
            In order to train the directions matrix $\mathbf{A}$ we minimize the following loss:
            \begin{equation}
            \begin{split}
                \mathcal{L}_{\mathbf{A}} = 
                    \lambda_{r}\mathcal{L}_{r}
                    + \lambda_{id}\mathcal{L}_{id} 
                    + \lambda_{per}\mathcal{L}_{per} + \\
                     \lambda_{pix}\mathcal{L}_{pix}
                    + \lambda_{style}\mathcal{L}_{style},
             \end{split}
            \end{equation}
            where $\mathcal{L}_{r}$, $\mathcal{L}_{id}$, $\mathcal{L}_{per}$, $\mathcal{L}_{pix}$, and $\mathcal{L}_{style}$ denote respectively the reenactment loss defined in Sect.~\ref{ssec:latent}, the identity, the perceptual, the pixel-wise, and the style losses calculated between the reenacted $\mathbf{I}_r$ and the target images $\mathbf{I}_t$. 
            
            \begin{figure}[t]
            \begin{center}
            \includegraphics[width=0.9\linewidth]{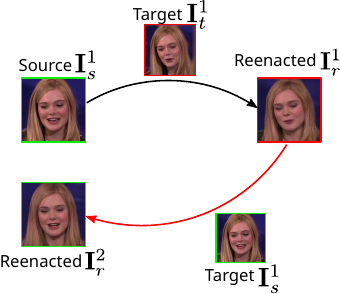}
            \end{center}
            \caption{\textbf{Cycle loss:} Given a pair of source ($\mathbf{I}_s^1$) and target ($\mathbf{I}_t^1$) images, we calculate the corresponding reenacted image $\mathbf{I}_r^1$. We then use this image as source and as target the source image from the first image pair and we calculate the second reenacted image $\mathbf{I}_r^2$, which is imposed to be similar with $\mathbf{I}_s^1$. }
            \label{fig:cycle_loss}
            \end{figure}
            
            Moreover, to further improve the reenactment results we propose an additional \textit{cycle loss} term $\mathcal{L}_{cycle}$~\cite{sanchez2020recurrent,bounareli2023stylemask}. Specifically, as shown in Fig.~\ref{fig:cycle_loss}, given an image pair of a source ($\mathbf{I}_s^1$) and a target ($\mathbf{I}_t^1$) images, we calculate the corresponding reenacted image $\mathbf{I}_r^1 \equiv \mathbf{I}_t^1$. Having as source image the reenacted image $\mathbf{I}_r^1$ and as target the source image $\mathbf{I}_s^1$, we calculate a new reenacted image $\mathbf{I}_r^2$ that is imposed to be similar to $\mathbf{I}_s^1$. Consequently, we calculate all reconstruction losses, i.e. $\mathcal{L}_{id}$, $\mathcal{L}_{per}$, $\mathcal{L}_{pix}$, and $\mathcal{L}_{style}$, between the source image $\mathbf{I}_s^1$ and the reenacted image $\mathbf{I}_r^2$. In our ablation studies (Sect.~\ref{ssec:ablation}), we show that using the proposed cycle loss improves the face reenactment performance.

        % === Overall objective ===
        \subsubsection{Joint optimization objective}
        
            Overall, the objective of the joint optimization is as follows:
            \begin{equation}
                \mathcal{L} = \mathcal{L}_\mathbf{A} + \mathcal{L}_{E_w} + \mathcal{L}_{cycle}.
            \end{equation}
            We note that, in this training phase, we fine-tune the matrix $\mathbf{A}$ and the real image inversion encoder $\mathcal{E}_w$, trained as described in Sect.~\ref{ssec:real}. As demonstrated in Fig.~\ref{fig:inversion}, using the proposed joint training scheme (Joint Training) our method is able to reconstruct the identity details of the real faces without performing any optimization step. In Sect.~\ref{sec:exp}, we quantitatively demonstrate that our proposed method produces similar results on self reenactment with our method when optimizing the generator's weights. Nevertheless, on the more challenging tasks of cross-subject reenactment and on large head pose differences between the source and target faces, the joint training scheme outperforms our results with optimization, producing more realistic images with less visual artifacts.

    %%%%%%%%%%%%%%%%%%%%%%%%%%%%%%%%%%%%%%%%%%%%%%%%%%%%%%%%%%%%%%%%%%%%%%%%%%%%%%%%
    %%                       [ Feature space F refinement ]                       %%
    %%%%%%%%%%%%%%%%%%%%%%%%%%%%%%%%%%%%%%%%%%%%%%%%%%%%%%%%%%%%%%%%%%%%%%%%%%%%%%%%
   \subsection{Feature space $\mathcal{F}$ refinement}\label{ssec:feature_space}

        \begin{figure}[t]
            \centering
            \includegraphics[width=1.0\linewidth]{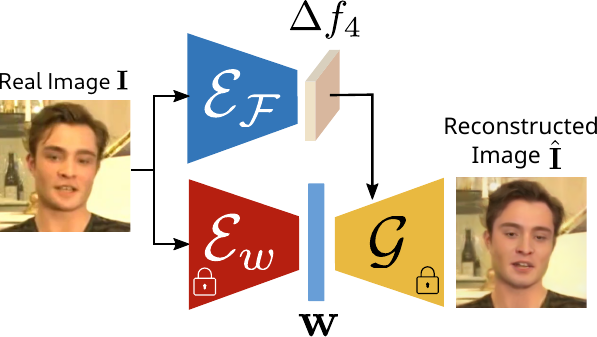}
            \caption{Training of feature space encoder $\mathcal{E}_{\mathcal{F}}$ in the real image inversion task. $\mathcal{E}_{\mathcal{F}}$ takes as input a real image and predicts the shift $\Delta f_4$ that updates the feature map $f_4$ of the $4^{th}$ feature layer of StyleGAN2's generator.} 
            \label{fig:feature_encoder}
        \end{figure}
        
        In this section, we propose an additional module for our face reenactment framework that refines the feature space $\mathcal{F}$ of the StyleGAN2's generator; taking advantage from its exceptional expressiveness (e.g., in terms of background, hair style/color, or hair accessories). In order to mitigate the limited editability of $\mathcal{F}$~\cite{parmar2022spatially,kang2021gan}, we propose a two-step training procedure, which we illustrate in Fig.~\ref{fig:feature_encoder}. Specifically, we first train a feature space encoder $\mathcal{E}_{\mathcal{F}}$, using the ResNet-18~\cite{he2016deep} architecture, in the task of real image inversion. $\mathcal{E}_{\mathcal{F}}$ takes as input a real image and predicts the shift $\Delta f_4$ that updates the feature map $f_4$ as:
        \begin{equation}\label{eq:feature_shift}
            \hat{f_4} = f_4 + \Delta f_4,
        \end{equation}
        where $f_4$ is the feature map calculated using the inverted latent code $\mathbf{w}$. The training objective in this step consists of the reconstruction losses, namely identity, perceptual, pixel-wise, and style, calculated between the reconstructed $\mathbf{\hat{I}}$ and the real images $\mathbf{I}$ as described in (Eq.~\ref{eq:recon_losses}). It is worth nothing that we only refine the $4^{th}$ feature layer of StyleGAN2's generator $\mathcal{G}$ that we found to be in particular beneficial to the face reenactment task, in contrast to later feature layers that, despite their capability in reconstructing almost perfectly the real images, they suffer from poor semantic editability (as shown by ~\cite{yao2022feature}).
        
\begin{figure}[t]
            \centering
            \includegraphics[width=1.0\linewidth]{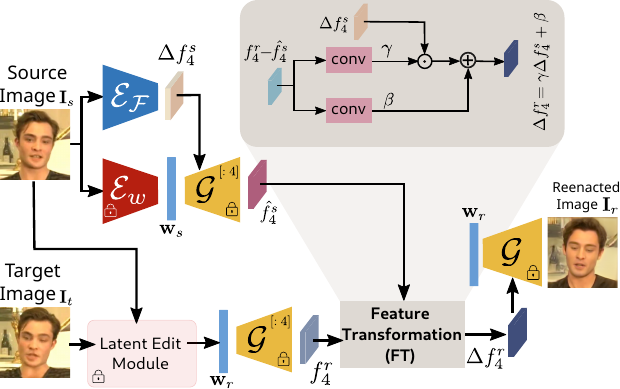}
            \caption{Training of the feature space encoder $\mathcal{E}_{\mathcal{F}}$ and the Feature Transformation (FT) module to efficiently refine the feature map $f_4^r$ of the reenacted images.} 
            \label{fig:feature_encoder_edit}
        \end{figure}

        As discussed above, using the updated feature map $\hat{f_4}$ to refine details on the edited images leads to visual artifacts. To address this, we propose a framework that efficiently learns to predict the updated feature map of the edited image $\hat{f_4^r}$ using the refined source feature map $\hat{f_4^s}$. We illustrate this in Fig.~\ref{fig:feature_encoder_edit}, where, given a source and a target image pair, we first calculate the reenacted latent code $\mathbf{w}_r$ as described in Sect.~\ref{ssec:joint_training}. We note that the directions matrix $\mathbf{A}$ and the real image inversion encoder $\mathcal{E}_w$ are frozen during training. Then, using the feature encoder $\mathcal{E}_{\mathcal{F}}$, we calculate the source refined feature map $\hat{f_4^s}$ using (\ref{eq:feature_shift}). In order to calculate the refined feature map of the reenacted image $\hat{f_4^r}$, we introduce the \textit{Feature Transformation (FT)} module, that takes as input the difference of the source refined feature map $\hat{f_4^s}$ and the reenacted feature map $f_4^r$, and outputs the shift $\Delta f_4^r$, which can be used to calculate the updated feature map $\hat{f_4^r}$ given by (\ref{eq:feature_shift}). As shown in Fig.~\ref{fig:feature_encoder_edit}, the proposed Feature Transformation (FT) module learns two modulation parameters, namely $\gamma$ and $\beta$, that efficiently transform the shift $\Delta f_4^s$ of the source feature map into the shift $\Delta f_4^r$ of the reenacted feature map as:
        \begin{equation}\label{eq:reenacted_feature}
            \Delta f_4^r = \gamma \odot \Delta f_4^s + \beta.
        \end{equation}
        As illustrated in Fig.~\ref{fig:feature_encoder_edit}, the FT module consists of two convolutional blocks with 2 convolutional layers each. We note that in this training step we train both the FT module and the feature space encoder $\mathcal{E}_{\mathcal{F}}$. Our training objective consists of the reconstruction losses, namely identity, perceptual, pixel-wise and style, calculated between the reenacted and the target images (described in detail in Sect.~\ref{ssec:joint_training}).

\begin{figure}[t]
            \centering
            \includegraphics[width=1.0\linewidth]{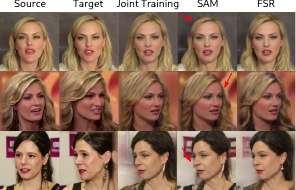}
            \caption{Face reenactment examples using only the $\mathcal{W}+$ latent space (``Joint Training''), SAM method~\cite{parmar2022spatially} and our proposed method for feature space refinement (``FSR'').} 
            \label{fig:feature_edit}
        \end{figure}

        % === Indicative results ===
        Finally, in Fig.~\ref{fig:feature_edit} we give some indicative results of the proposed reenactment variant of our method that learns to optimize the feature space $\mathcal{F}$  (``FSR'') in comparison to the variant of our method described in Sect.~\ref{ssec:joint_training} (``Joint Training'') and~\cite{parmar2022spatially} (``SAM''). We note that using the $\mathcal{W}+$ latent space (Joint Training / Sect.~\ref{ssec:joint_training}) leads to relatively faithful reconstruction performance, albeit, without being able to reconstruct every detail on the background or the hair styles. As we will show in the experimental section, qualitatively and quantitatively, but also in the conducted user study, such level of detail is crucial for the task of face reenactment. Similarly, SAM~\cite{parmar2022spatially} is able to better reconstruct the background however the reenacted images suffer from visual artifacts (marked with red arrows in Fig.~\ref{fig:feature_edit}) and, thus, look unrealistic, especially around the face area. By contrast, the proposed framework that learns to optimize the feature space $\mathcal{F}$ (``FSR'') leads to both notably more faithful face reenactment exhibiting less artifacts.
        
%%%%%%%%%%%%%%%%%%%%%%%%%%%%%%%%%%%%%%%%%%%%%%%%%%%%%%%%%%%%%%%%%%%%%%%%%%%%%%%%
%%                                                                            %%
%%                             [ Experiments ]                                %%
%%                                                                            %%
%%%%%%%%%%%%%%%%%%%%%%%%%%%%%%%%%%%%%%%%%%%%%%%%%%%%%%%%%%%%%%%%%%%%%%%%%%%%%%%%
\section{Experiments}\label{sec:exp}

    In this section, we present qualitative and quantitative results, along with a user study, in order to evaluate the proposed framework (all its variants) in the task of neural face reenactment and compare with several recent state-of-the-art (SOTA) approaches. The bulk of our results and comparisons, reported in Sect.~\ref{ssec:face_re}, are on self- and cross-person reenactment on the VoxCeleb1~\cite{Nagrani17} dataset. Comparisons with state-of-the-art on the VoxCeleb2~\cite{Chung18b} test set are provided in the appendices. Finally, in Sect.~\ref{ssec:ablation} we report ablation studies on the various design choices of our method and in Sect.~\ref{ssec:limitations} we discuss its limitations. 

    \paragraph{Implementation details} We fine-tune StyleGAN2 on the VoxCeleb1 dataset with $256\times256$ image resolution and we train the e4e encoder of~\cite{tov2021designing} for real image inversion. The 3D shape model we use (i.e., the Net3D module shown in Figs.~\ref{fig:architecture},~\ref{fig:architecture_joint}) is DECA~\cite{feng2020deca}. For our training procedure described in Sects.~\ref{ssec:latent}~\ref{ssec:real}, and~\ref{ssec:video}, we only learn the directions matrix $\mathbf{A}\in \mathbb{R}^{(N_l\times512) \times k}$ where $k=3+m_e, m_e = 12$ and $N_l=8$. We train three matrices of directions: (i) the first one is on synthetically generated images (Sect.~\ref{ssec:latent}), (ii) the second one is on mixed real and synthetic data (Sect.~\ref{ssec:real}), and (iii) the third one is fine-tuning (ii) on paired data (Sect.~\ref{ssec:video}). Additionally, on the proposed joint training scheme (Sect.~\ref{ssec:joint_training}), we fine-tune both the directions matrix $\mathbf{A}$ and the real image inversion encoder $\mathcal{E}_w$. Finally, in the feature space refinement variant (Sect.~\ref{ssec:feature_space}) we train both the feature space encoder $\mathcal{E}_{\mathcal{F}}$ and the propose Feature Transformation (FT) module. It is worth noting that during the first and second training phases, we perform cross-subject training, i.e., the source and target faces have different identities. This approach enables our model to generalize effectively across various identities, resulting in improved performance on the challenging task of cross-subject reenactment. On the rest training phases we perform self reenactment, where the source and target faces are sampled from the same video. For training, we used the Adam optimizer~\cite{kingma2014adam} with constant learning rate $10^{-4}$. All models are implemented in PyTorch~\cite{paszke2017automatic}.

    \subsection{Comparison with state-of-the-art on VoxCeleb}\label{ssec:face_re}

        In this section, we compare the performance of our method against the state-of-the-art in face reenactment on VoxCeleb1~\cite{Nagrani17}. We conduct two types of experiments, namely self- and cross-person reenactment. For evaluation purposes, we use both the video data provided by~\cite{zakharov2019few} and the original test-set of VoxCeleb1. We note that there is no overlap between the train and test identities and videos. We compare our method quantitatively and qualitatively with nine methods: X2Face~\cite{wiles2018x2face}, FOMM~\cite{siarohin2019first}, Fast bi-layer~\cite{zakharov2020fast}, Neural-Head~\cite{burkov2020neural}, LSR~\cite{meshry2021learned}, PIR~\cite{ren2021pirenderer}, HeadGAN~\cite{doukas2020headgan}, Dual~\cite{hsu2022dual} and Face2Face~\cite{yang2022face2face}. For X2Face~\cite{wiles2018x2face}, FOMM~\cite{siarohin2019first}, PIR~\cite{ren2021pirenderer}, HeadGAN~\cite{doukas2020headgan} and Face2Face~\cite{yang2022face2face}, we use the pre-trained (by the authors) model on VoxCeleb1. For Fast bi-layer~\cite{zakharov2020fast}, Neural-Head~\cite{burkov2020neural} and LSR~\cite{meshry2021learned} we also use the pre-trained (by the authors) models on VoxCeleb2~\cite{Chung18b}. Regarding Dual~\cite{hsu2022dual}, we use the pre-trained by the authors model on both VoxCeleb~\cite{Nagrani17, Chung18b} and MPIE~\cite{gross2010multi} datasets. For fair comparison with the methods of Neural-Head~\cite{burkov2020neural}, LSR~\cite{meshry2021learned} and Dual~\cite{hsu2022dual}, we evaluate their model under the one-shot setting. We note that we will be referring to our method that optimizes the generator's weights during inference as Latent Optimization Reenactment (LOR), whereas LOR+ will be referring to our final model with joint training and feature space refinement. We note that in the LOR+ model, we do not optimize the generator weights.

        \subsubsection{Quantitative comparisons}\label{subsubsec:quanti_comp}

            We report eight different metrics. We compute the Learned Perceptual Image Path Similarity (LPIPS)~\cite{zhang2018unreasonable} to measure the perceptual similarities, and to quantify identity preservation we compute the cosine similarity (CSIM) of ArcFace~\cite{deng2019arcface} features. Moreover, we measure the quality of the reenacted images using the Fr\'echet-Inception Distance (FID) metric~\cite{heusel2017gans}, while we also report the Fr\'echet Video Distance (FVD)~\cite{unterthiner2018towards,skorokhodov2022stylegan} metric that measures both the video quality and the temporal consistency of the generated videos. To quantify the head pose/expression transfer, we calculate the normalized mean error (NME) between the predicted landmarks in the reenacted and target images. We use~\cite{bulat2017far} for landmark estimation, and we calculate the NME by normalizing it with the square root of the ground truth face bounding box and scaled by a factor of $10^3$. We further evaluate the head pose transfer by calculating the average $\mathcal{L}_1$ distance of the head pose orientation (Average Rotation Distance, ARD) in degrees, and the expression transfer by calculating the average $\mathcal{L}_1$ distance of the expression coefficients $\mathbf{p}_{e}$ (Average Expression Distance, AED) and the Action Units Hamming distance (AU-H) computed as in~\cite{doukas2020headgan}.

            \begin{table*}[t]
            \caption{Quantitative results on self-reenactment. The results are reported on the combined original test set of VoxCeleb1~\cite{Nagrani17} and the test set released by~\cite{zakharov2019few}. For CSIM metric, higher is better ($\uparrow$), while in all other metrics lower is better ($\downarrow$).}\label{table:self_reenactment_metrics}
            \begin{center}
            \begin{tabular}{|c|c|c|c|c|c|c|c|c|}
                \hline
                Method & CSIM & LPIPS & FID & FVD & NME & ARD& AED & AU-H \\
                \hline
                X2Face~\cite{wiles2018x2face} & 0.70 & 0.13  & \underline{35.5} & 490 & 17.8 & 1.5 & 0.90 & \underline{0.22} \\
                FOMM~\cite{siarohin2019first} & 0.65 & 0.14 & 35.6 & 523 & 34.1 & 5.0 & 1.30 & 0.28 \\
                Fast Bi-layer~\cite{zakharov2020fast} & 0.64 & 0.23  & 52.8 & 706 & \textbf{13.2} & 1.1 & 0.80 & \textbf{0.21} \\
                Neural-Head~\cite{burkov2020neural} & 0.40 & 0.22  & 98.4 & 617 & 15.5 & 1.3 & 0.90 & 0.23 \\
                LSR~\cite{meshry2021learned} & 0.59 & 0.13 & 45.7 & 484 & 17.8 & \underline{1.0} & 0.75 & \underline{0.22}\\
                PIR~\cite{ren2021pirenderer} & \underline{0.71} & 0.12 & 57.2 & 545 & 18.2 & 1.8 & 0.94 & 0.24 \\
                HeadGAN~\cite{doukas2020headgan} & 0.68 & 0.13 & 52.5 & 518 & 15.6 & 1.8 & 1.30 & 0.26 \\ 
                Dual~\cite{hsu2022dual} & 0.26 & 0.21  & 75.0  & 602 & 35.0 & 3.7  & 1.20  & 0.27\\
                Face2Face~\cite{yang2022face2face} & \textbf{0.72} & 0.12 & 55.3 & 682 & 16.0 & 1.5 & 0.93 & 0.24 \\
                LOR (Ours) & 0.66 & \underline{0.11} & \textbf{35.0} & \textbf{400} & 14.1 & 1.1 & \underline{0.68} & \textbf{0.21} \\
                LOR+ (Ours) & 0.67 & \textbf{0.10} & 36.0  & \underline{440}  & \underline{13.6} & \textbf{0.7} & \textbf{0.60} & \textbf{0.21} \\ 
                \hline
            \end{tabular}
            \end{center}
            \end{table*}

            \begin{table*}[t]
            \begin{center}
            \caption{Quantitative comparisons on the benchmark set (Benchmark-L ) with image pairs from VoxCeleb1 dataset, where the average head pose distance is larger than $10^{\circ}$. For CSIM metric, higher is better ($\uparrow$), while in all other metrics lower is better ($\downarrow$).}\label{table:self_reenactment_large_metrics}
            \begin{tabular}{|c|c|c|c|c|c|c|}
            \hline
            Method & CSIM & LPIPS & FID & ARD & AED & AU-H \\
             \hline
            X2Face~\cite{wiles2018x2face} & \underline{0.60} & 0.14 & 57.4 & 1.8  & 1.1 & 0.25\\
            FOMM~\cite{siarohin2019first} & \underline{0.60} & 0.15 & 65.2 & 2.2 & 1.1 & 0.25\\
            Fast Bi-layer~\cite{zakharov2020fast} & 0.58 & 0.20 & 96.2 & \underline{1.2} & \underline{0.8} & \textbf{0.22}\\
            Neural-Head~\cite{burkov2020neural} & 0.40 & 0.18 & 94.2 & \underline{1.2} & 0.9 & \underline{0.23}\\
            LSR~\cite{meshry2021learned} & 0.55 & \textbf{0.12} & 56.0 & \underline{1.2} & \underline{0.8} & \underline{0.23}\\
            PIR~\cite{ren2021pirenderer} & 0.57 & 0.15 & 67.6 & 2.4 & 1.4 & 0.25\\
            HeadGAN~\cite{doukas2020headgan} & 0.38 & 0.26 & 66.2 & 3.6 & 1.4 & 0.29 \\
            Dual~\cite{hsu2022dual} & 0.25 & 0.22 & 83.6 & 4.0  & 1.3 & 0.28\\
            Face2Face~\cite{yang2022face2face} & 0.47 & 0.28 & \textbf{35.3} & 1.6 & 1.2 & 0.27\\
            LOR (Ours) & 0.51 & \underline{0.13} & 47.0 & 1.6 & \underline{0.8} & 0.35\\
            LOR+ (Ours)  & \textbf{0.62} & \textbf{0.12} & \underline{46.7} & \textbf{0.8}  &\textbf{0.6} & \textbf{0.22}\\
            \hline
            \end{tabular}
            \end{center}
            \end{table*}

            \begin{table*}[t]
            \begin{center}
            \caption{Quantitative results on cross-subject reenactment. For CSIM metric, higher is better ($\uparrow$), while in all other metrics lower is better ($\downarrow$).}\label{table:cross_reenactment_metrics}
            \begin{tabular}{|c|c|c|c|c|c|c|c|c|c|c|c|}
            \hline
             Method & CSIM  & FID  & ARD  & AED & AU-H \\
             \hline
            X2Face~\cite{wiles2018x2face} &  0.57 & 89.0 & 2.2 & 1.5 & 0.31\\
            FOMM~\cite{siarohin2019first} & 0.73 & 116.3 & 7.7 & 2.0 & 0.41\\
            Fast Bi-layer~\cite{zakharov2020fast} & 0.48 & 116.0 & 1.5 & 1.3 & \textbf{0.29}\\
            Neural-Head~\cite{burkov2020neural}  & 0.36 & 124.1 & 1.7 & 1.6 & \underline{0.30}\\
            LSR~\cite{meshry2021learned}  & 0.50 & \underline{84.2} & 1.4 & 1.2 & \underline{0.30}\\
            PIR~\cite{ren2021pirenderer} & 0.62 & 110.5 & 2.2 & 1.4  & 0.33\\
             HeadGAN~\cite{doukas2020headgan} & \underline{0.75} & 122.2 & 2.1 & 1.7 & 0.33  \\
            Dual~\cite{hsu2022dual} & 0.22 & 107.0 & 3.5 & 1.5 & 0.33\\
            Face2Face~\cite{yang2022face2face} & \textbf{0.76} & 124.2 & 1.8 & 1.5 & 0.32\\
            LOR (Ours) & 0.63 & 86.2 &  \underline{1.2} & \underline{1.0} & 0.31\\
            LOR+ (Ours) & 0.68 & \textbf{78.4} &  \textbf{0.7} & \textbf{0.8} & \underline{0.30} \\
            \hline
            \end{tabular}
            \end{center}
            \end{table*}

\begin{figure*}[t]
            \begin{center}
            {\includegraphics[width=1.0\textwidth]{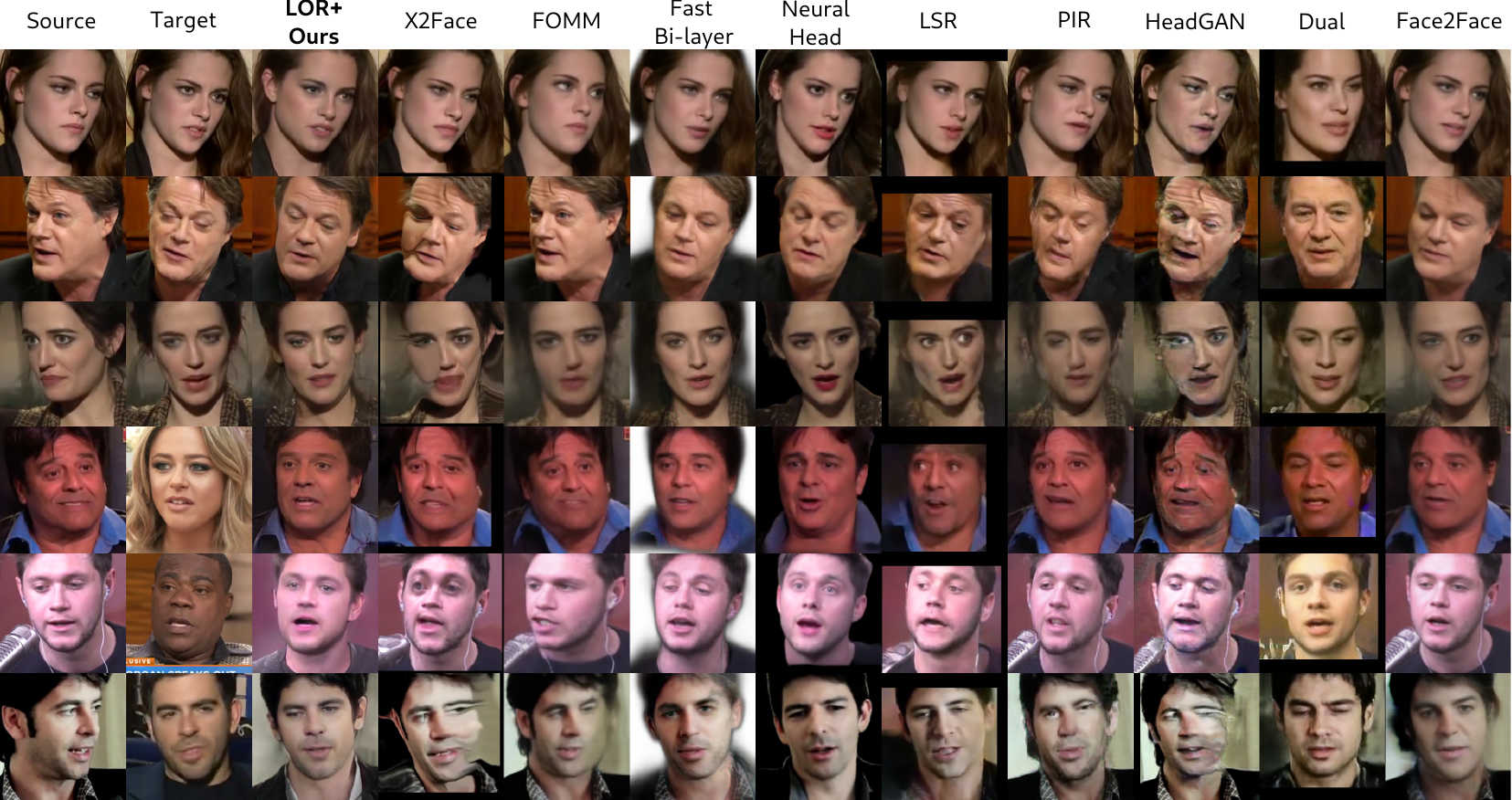}}
            \end{center}
              \caption{Qualitative results and comparisons for self (top three rows) and cross-subject reenactment (last three rows) on VoxCeleb1. The first and second columns show the source and target faces. Our method preserves the appearance and identity characteristics (e.g., face shape) of the source face significantly better and also faithfully transfer the target head pose/expression without producing visual artifacts.}
            \label{fig:comparisons_all}
            \end{figure*}

            In Table~\ref{table:self_reenactment_metrics} we report quantitative results on self reenactment, using the original test set of VoxCeleb1~\cite{Nagrani17} and the test set provided by~\cite{zakharov2019few}. Additionally, in Table~\ref{table:self_reenactment_large_metrics} we report results on a more challenging condition on self reenactment where the source and target faces have large head pose difference. Specifically, we randomly selected from the test set of VoxCeleb1 1,000 image pairs with head pose distance larger than $10^{\circ}$. The head pose distance is calculated as the average of the absolute differences of the three Euler angles (i.e., yaw, pitch, and roll) between the source and target faces. In the appendices (Sect.~\ref{subsec:bechmark}), we provide additional details regarding our benchmark dataset. We note that in self reenactment, all metrics are calculated between the reenacted and the target faces. As shown in Table~\ref{table:self_reenactment_metrics}, the warping-based methods, namely X2Face, PIR, HeadGAN and Face2Face have high values on CSIM, however we argue that this is due to their warping-based technique which enables better reconstruction of the background and other identity characteristics. Importantly, these results are accompanied by poor quantitative and qualitative results when there is a significant change on the head pose (e.g., see Fig.~\ref{fig:comparisons_all} and Table~\ref{table:self_reenactment_large_metrics}). Additionally, regarding head pose/expression transfer, our method (LOR+) achieves similar results on NME with Fast Bi-layer~\cite{zakharov2020fast}, while on ARD and AED metrics we outperform all methods. Finally, our results on FID and FVD metrics confirm that the quality of our generated videos resembles the quality of VoxCeleb dataset. Nevertheless, our method (LOR+) on the challenging condition with large head pose differences between the source and target faces (Table~\ref{table:self_reenactment_large_metrics}) outperforms all methods. 

            Cross-subject reenactment is more challenging compared to self reenactment, as source and target faces have different identities, and in this case it is important to maintain the source identity characteristics without transferring the target ones. In Table~\ref{table:cross_reenactment_metrics}, we report the quantitative results for cross-subject reenactment, where we randomly select 200 video pairs from the small test set of~\cite{zakharov2019few}. In this task, CSIM metric is calculated between the source and the reenacted faces while ARD, AED and AU-H metrics between the target and the reenacted faces. As depicted in Table~\ref{table:cross_reenactment_metrics}, our method (LOR+) achieves the best results on head pose and expression transfer, while we achieve high score in CSIM metric. It is worth noting that the high CSIM value for FOMM, HeadGAN and Face2Face is not accompanied by good qualitative results as shown in Figs.~\ref{fig:comparisons_all} and~\ref{fig:cross_vox1}, where in most cases, those methods are not able to generate realistic images.

            To further evaluate the performance of reenactment methods we conduct a user study, where we ask 30 users to select the method that best reenacts the source frame on self and cross-subject reenactment tasks. For the purposes of the user study we utilise only our final model (LOR+). The results are reported in Table~\ref{table:user_study} and as shown our method is the most preferable (by a large margin -- 52.1\% versus the 19.2\% second best method), which also validates our quantitative results.

            Additionally, in Table~\ref{table:inference_time} we report comparisons on inference time required to generate a video of $200$ frames. As shown, X2Face~\cite{wiles2018x2face} and FOMM~\cite{siarohin2019first} are the fastest methods, however their overall performance (quantitative and qualitative results) is unsatisfactory (i.e., visual artifacts). Nevertheless, our proposed method (LOR+) is able to generate compelling reenacted images, while also being competitive in terms of inference time. Notably, our final model (LOR+) outperforms our model that requires the optimization step (LOR), which is a time consuming operation.
            
            \begin{table}[t]
            \caption{ Quantitative comparisons on inference time required to generate a video of $200$ frames.}\label{table:inference_time}
            \begin{tabular}{|c|c|}
                \hline
                Method & Inf. time (sec)  \\
                \hline
                X2Face~\cite{wiles2018x2face} & \textbf{11.0} \\
                FOMM~\cite{siarohin2019first} & \textbf{11.0}\\
                Fast Bi-layer~\cite{zakharov2020fast} & 61.0\\
                Neural-Head~\cite{burkov2020neural} & 115.0\\        
                LSR~\cite{meshry2021learned} & 110.0\\
                PIR~\cite{ren2021pirenderer} & 54.0\\    
                HeadGAN~\cite{doukas2020headgan} & 84 \\
                Dual~\cite{hsu2022dual} & \underline{27.0}\\
                Face2Face~\cite{yang2022face2face} & 146 \\
                LOR (Ours) & 40.0\\
                LOR+ (Ours) & \underline{27.0}\\ 
                \hline
            \end{tabular}
            \end{table}

        \subsubsection{Qualitative comparisons}
            
            Quantitative comparisons alone are insufficient to capture the quality of reenactment. Hence, we opt for qualitative visual comparisons \textit{in multiple ways}: (a) results in Fig.~\ref{fig:comparisons_all}, (b) in the appendices, we provide more results in self and cross-subject reenactment both on VoxCeleb1 and VoxCeleb2 datasets (Figs.~\ref{fig:comparisons_large},~\ref{fig:self_vox1},~\ref{fig:cross_vox1},~\ref{fig:cross_vox1_2},~\ref{fig:vox2_results}), and (c) we also provide a supplementary video with self and cross-subject reeenactment results from VoxCeleb1 and VoxCeleb2 datasets. As we can see from Fig.~\ref{fig:comparisons_all} and the videos provided in the supplementary material, our method provides, for the majority of videos, the highest reenactment quality including accurate transfer of head pose and expression and, significantly enhanced identity preservation compared to all other methods. Importantly, one great advantage of our method on cross-subject reenactment, as shown in Fig.~\ref{fig:comparisons_all}, is that it is able to reenact the source face with minimal identity leakage (e.g facial shape) from the target face, in contrast to landmark-based methods such as Fast Bi-layer~\cite{zakharov2020fast}. Finally, to show that our method is able to generalise well on other facial video datasets, we provide additional results on the FaceForensics~\cite{roessler2018faceforensics} and 300-VW~\cite{shen2015first} datasets in the appendices (Fig.~\ref{fig:faceforensics}).

    \subsection{Ablation studies}\label{ssec:ablation}
        
        In this section, we perform several ablation tests to (a) assess the different variants of our method, i.e., the optimization of generator $\mathcal{G}$ during inference (Sect.~\ref{ssec:video}), the proposed joint training scheme (Sect.~\ref{ssec:joint_training}) and the refinement of the feature space (Sect.~\ref{ssec:feature_space}), (b) measure the impact of the identity and perceptual losses, and the additional shape losses for the eyes and mouth (Sect.~\ref{ssec:latent}), (c) validate our trained models on synthetic, mixed and paired images, and (d) measure the impact of the style and cycle losses (introduced in Sect.~\ref{ssec:joint_training}).
        
        \begin{table*}[!t]
        \begin{center}
        \caption{Quantitative results of the various models of our work on self reenactment (SR), self reenactment with image pairs that have large head pose difference (SR - large head pose) and on cross-subject reenactment (CR).}
        \label{table:diff_components}
        \begin{tabular}{|l|c|c|c|c|c|c|c|c|c|c|}
        \hline
        \multirow{2}{*}{Method} & \multicolumn{4}{c|}{SR} & \multicolumn{3}{c|}{SR-large head pose} & \multicolumn{3}{c|}{CR} \\
        \hhline{~|----------}
         & CSIM & LPIPS & ARD & AED & CSIM &  ARD & AED & CSIM & ARD & AED \\
         \hline
        LOR \textit{w/o opt.} & 0.37 & 0.12 & 1.4 &  0.90 & 0.34 & 1.7 & 0.9 & 0.43 & 1.5 & \underline{1.0}  \\
        
        LOR \textit{w/ opt.} & \underline{0.66} & \underline{0.11} & 1.1 & \underline{0.68} & 0.51 & 1.6 & \underline{0.8} & 0.63 &  \underline{1.2} & \underline{1.0} \\
        
        LOR+ \textit{w/o FSR} & \underline{0.66} & \underline{0.11} & \underline{0.8}  & \textbf{0.60} &  \underline{0.60} & \underline{0.9} & \textbf{0.6} & \underline{0.67} & \textbf{0.7} & \textbf{0.8} \\
        
        LOR+ \textit{w/ FSR} & \textbf{0.67} & \textbf{0.10} & \textbf{0.7} & \textbf{0.60}  & \textbf{0.62} & \textbf{0.8} & \textbf{0.6} & \textbf{0.68} & \textbf{0.7} & \textbf{0.8}\\
        \hline
        \end{tabular}
        \end{center}
        \end{table*}
        
\begin{figure}[t]
        \begin{center}
        {\includegraphics[width=1.0\linewidth]{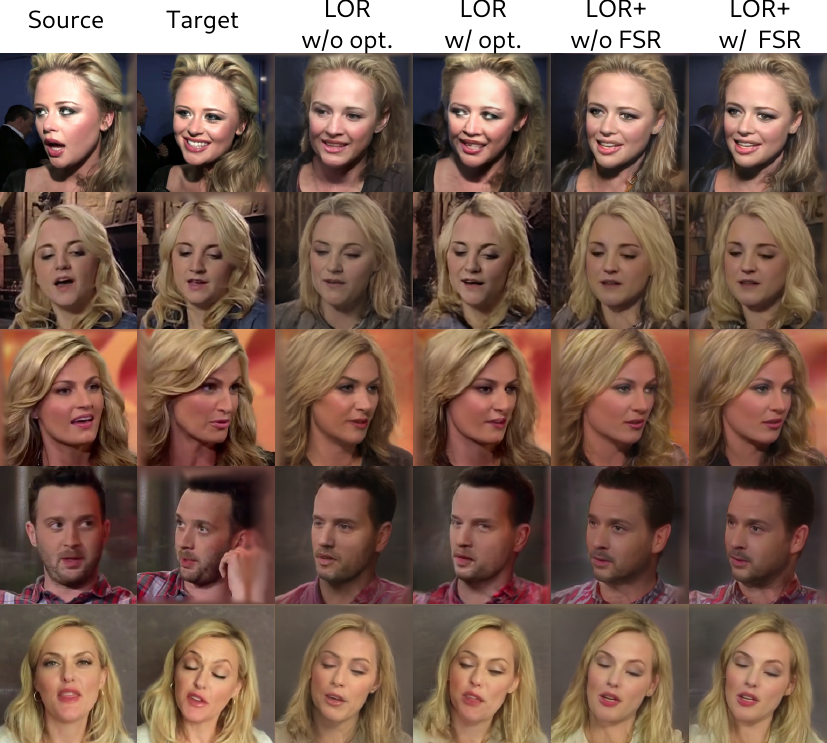}}
        \end{center}
          \caption{Qualitative comparisons of the various models of our work on self reenactment.}
        \label{fig:comparisons_mine}
        \end{figure}

        For (a), we report results of our method on self and cross-subject reenactment, with our model (LOR) described in Sect.~\ref{ssec:video} without performing optimization (w/o opt.) and with optimization (w/ opt.) of the generator $\mathcal{G}$ during inference. We also report results of our final model (LOR+) without the additional feature space refinement (FSR) (Sect.~\ref{ssec:joint_training}) and with feature space refinement (Sect.~\ref{ssec:feature_space}). As shown in Table~\ref{table:diff_components}, the optimization of $\mathcal{G}$ during inference improves our results (as expected) especially regarding the identity preservation (CSIM) compared to our model without performing optimization. Nevertheless, our proposed joint training scheme (LOR+ \textit{w/o FSR}) achieves the same results on image reconstruction metrics (CSIM and LPIPS), and improves our results on head pose/expression transfer (ARD, AED) without performing any optimization of the generator. Additionally, the proposed refinement on the feature space of StyleGAN2 (LOR+ \textit{w/ FSR}) improves all quantitative results. It is worth mentioning that the new proposed components (Joint Training and Feature Space Refinement) compared to our previous work~\cite{bounareli2022finding} improve our results especially on the challenging tasks of self reenactment with large head pose differences between the source and target faces and on cross-subject reenactment. Fig.~\ref{fig:comparisons_mine} illustrates results on self reenactment using the above described models. As shown LOR without optimization cannot accurately reconstruct the identity of the source face, while with optimization the identity details are better reconstructed but the reenacted images contain noticeable visual artifacts. On the contrary, the proposed joint training scheme (LOR+ w/o FSR) is able to accurately reconstruct the identity of the source faces and produce artifact-free images without performing any subject fine-tuning. Finally, the proposed feature space refinement (LOR+ w/ FSR) improves our qualitative results by producing more realistic images (i.e., better background and hair style reconstruction).

\begin{figure}[h]
            \begin{center}
            \includegraphics[width=1.0\linewidth]{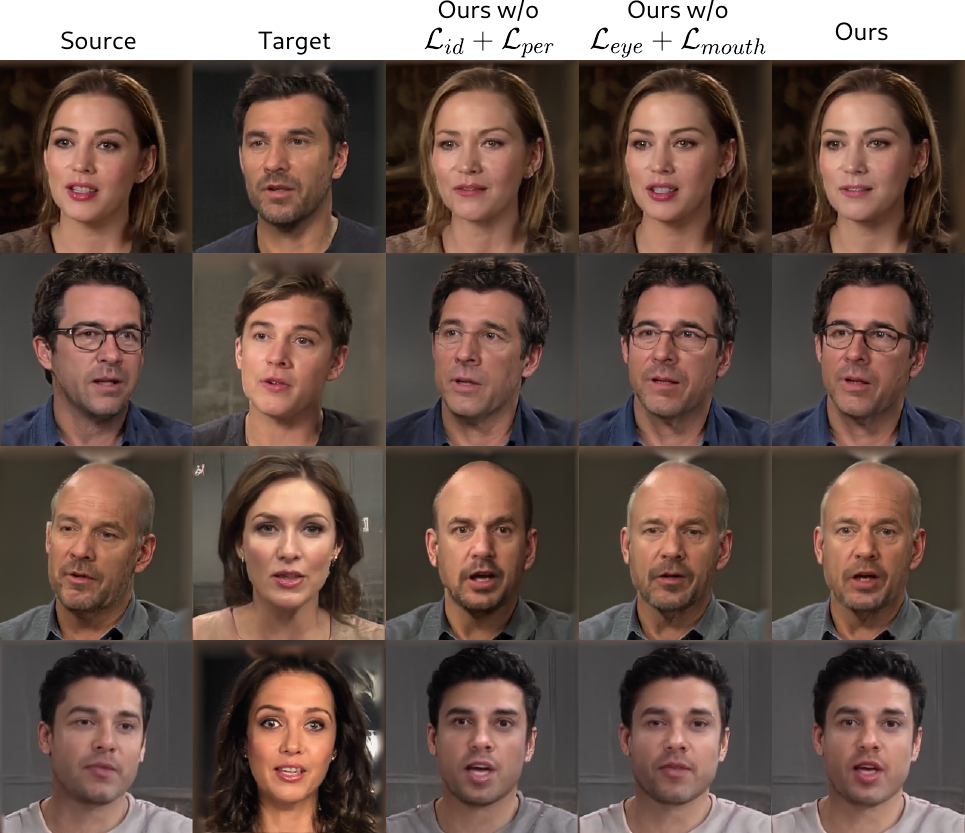}
            \end{center}
            \caption{Qualitative comparisons on the impact of the identity $\mathcal{L}_{id}$ and perceptual $\mathcal{L}_{per}$ losses, and on the impact of eye $\mathcal{L}_{eye}$ and mouth $\mathcal{L}_{mouth}$ losses.}
            \label{fig:ablation_table_5}
        \end{figure}

        For (b), we perform experiments on synthetic images with and without the identity and perceptual losses. To evaluate the models, we randomly generate $5K$ pairs of synthetic images (source and target) and reenact the source image with the head pose and expression of the target. As shown in Table~\ref{table:ablation_id}, the incorporation of the identity and perceptual losses is crucial to isolate the latent space directions that strictly control the head pose and expression characteristics without affecting the identity of the source face. In a similar fashion, in Table~\ref{table:ablation_id}, we show the impact of the additional shape losses, namely the eye $\mathcal{L}_{eye}$ and mouth $\mathcal{L}_{mouth}$ losses. As shown, omitting these losses leads to higher head pose and expression error. The impact of those losses is also obvious on our qualitative comparisons in Fig.~\ref{fig:ablation_table_5}. As shown, when we exclude the identity and perceptual losses from the training process, the generated images lack several appearance details, while omitting the eye and mouth losses leads to less accurate facial expression transfer.
        
        \begin{table}[t]
        \caption{Ablation study on the impact of the identity $\mathcal{L}_{id}$ and perceptual $\mathcal{L}_{per}$ losses, and on the impact of eye $\mathcal{L}_{eye}$ and mouth $\mathcal{L}_{mouth}$ losses. CSIM is calculated between the source and the reenacted images which are on different head pose and expression.}
        \begin{tabular}{|c|c|c|c|}
        \hline
        Method & CSIM  & ARD  & AED \\
        \hline
        Ours w/ $\mathcal{L}_{id} +\mathcal{L}_{per}$  & \textbf{0.52} & 2.4  & 1.2  \\
        Ours w/o $\mathcal{L}_{id} + \mathcal{L}_{per}$ & 0.42 & 2.5 & 1.2\\
        \hline
        Ours w/ $\mathcal{L}_{eye} +\mathcal{L}_{mouth}$ & 0.52 & \textbf{2.4} & \textbf{1.2} \\
        Ours w/o $\mathcal{L}_{eye} + \mathcal{L}_{mouth}$  & 0.52 & 2.6 & 1.5\\
        \hline
        \end{tabular}
        \label{table:ablation_id}
        \end{table}

\begin{figure}[h]
            \begin{center}
            \includegraphics[width=1.0\linewidth]{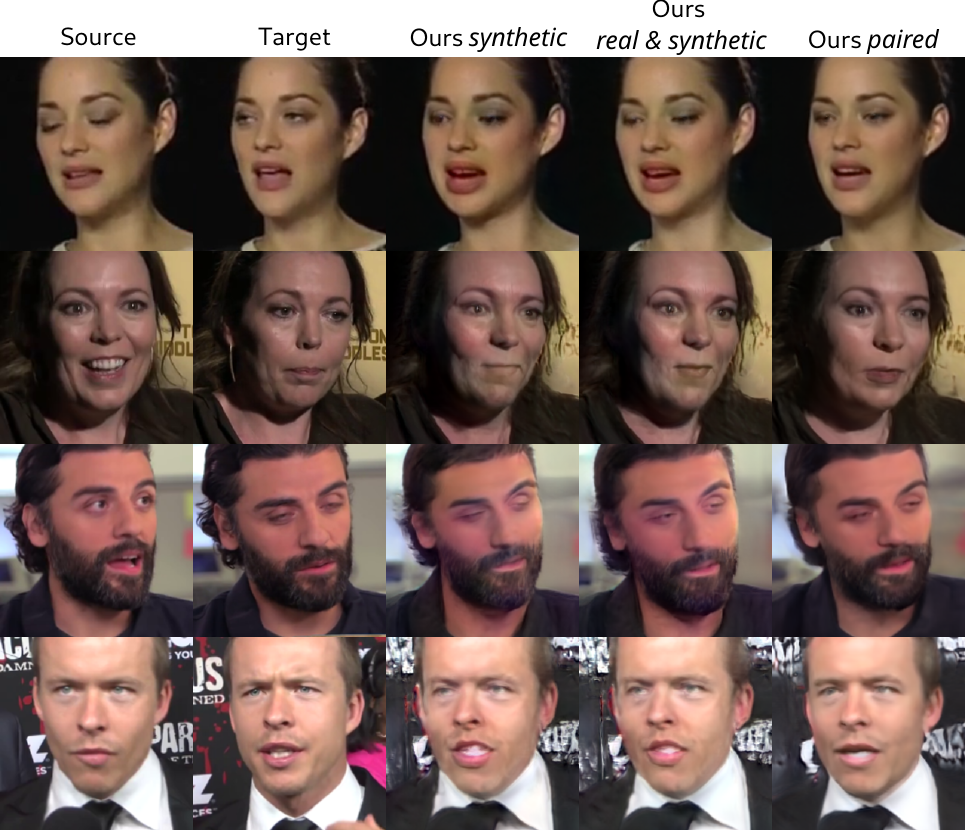}
            \end{center}
            \caption{Qualitative results of the three different models trained on synthetic images, on both synthetic and real images and on paired data.}
            \label{fig:ablation_table_6}
        \end{figure}

        For (c), we evaluate the three different training schemes, namely synthetic only (Sect.~\ref{ssec:latent}), mixed synthetic-real (Sect.~\ref{ssec:real}), and mixed synthetic-real fine-tuned with paired data (Sect.~\ref{ssec:video}) for self-reenactment. The results, reported in Table~\ref{table:diff_models} and in Fig.~\ref{fig:ablation_table_6}, illustrate the impact of each of these training schemes with the one using paired data providing the best results as expected. As shown in Fig.~\ref{fig:ablation_table_6}, our final model trained with paired data produces more realistic images with less artifacts.
        
        \begin{table}[t]
        \caption{Ablation studies on self-reenactment using three different models: (a) trained on synthetic images, (b) trained on both synthetic and real images, and (c) fine-tuned on paired data.}
        \label{table:diff_models}
        \begin{tabular}{|c|c|c|c|}
        \hline
        Method & CSIM & ARD  & AED  \\
        \hline
        Ours \textit{synthetic}  & 0.60 & 1.7 & 1.1\\
        Ours \textit{real} \& \textit{synthetic} & 0.63 & 1.6 & 1.1\\
        Ours \textit{paired} & \textbf{0.66} & \textbf{1.1} & \textbf{0.8}\\
        \hline
        \end{tabular}
        \end{table}

\begin{figure}[h]
            \begin{center}
            \includegraphics[width=1.0\linewidth]{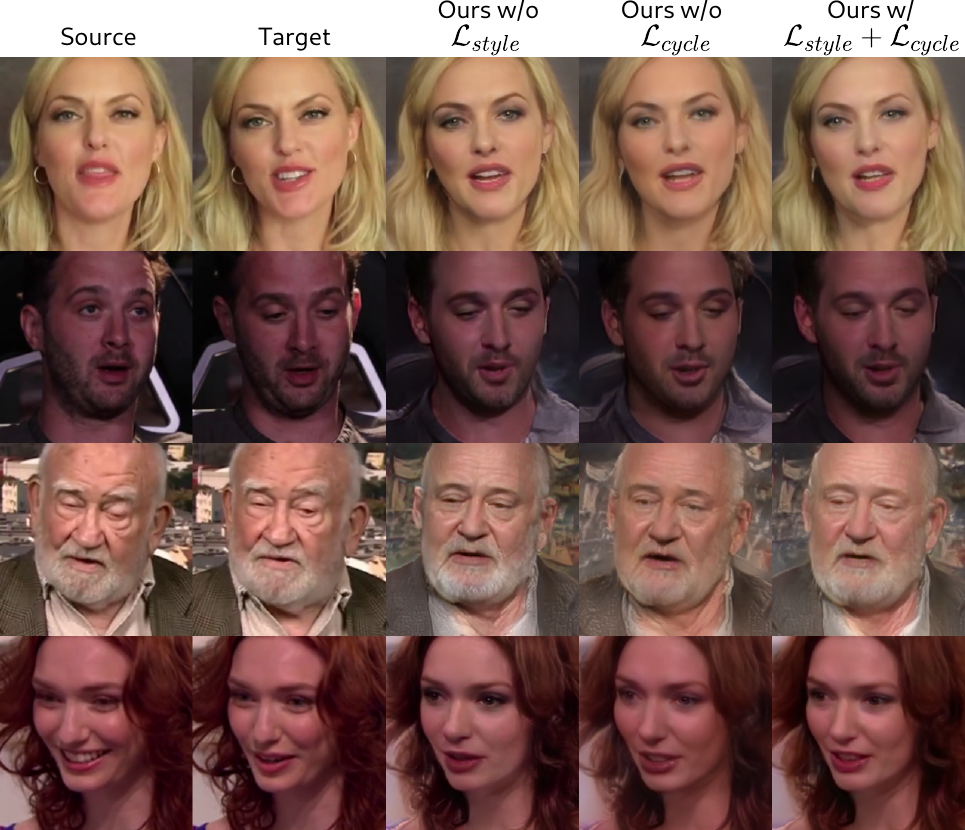}
            \end{center}
            \caption{Qualitative comparisons on the impact of the style $\mathcal{L}_{style}$ and cycle  $\mathcal{L}_{cycle}$ losses.}
            \label{fig:ablation_table_7}
        \end{figure}

        Finally, for (d) we perform experiments on self reenactment using our model with joint training scheme, without using the style loss $\mathcal{L}_{style}$ and without the cycle loss $\mathcal{L}_{cycle}$. As shown in Table~\ref{table:ablation_farl} our final model with both those losses has better results both on identity preservation and on head pose/expression transfer. Additionally, as illustrated in Fig.~\ref{fig:ablation_table_7}, our final model using both the style and the cycle loss has improved results in terms of identity/appearance preservation.
        
        \begin{table}[t]
        \caption{Ablation study on the impact of style $\mathcal{L}_{style}$ and cycle  $\mathcal{L}_{cycle}$ losses.}
        \label{table:ablation_farl}
        \begin{tabular}{|c|c|c|c|}
        \hline
        Method & CSIM  & ARD  & AED \\
        \hline
        Ours w/o $\mathcal{L}_{style}$ & 0.64 &  0.9  & 0.7  \\
        Ours w/o $\mathcal{L}_{cycle}$ & 0.62 & 1.0 & 0.7 \\
        Ours w/ $\mathcal{L}_{style} +\mathcal{L}_{cycle}$  & \textbf{0.66} & \textbf{0.8}  & \textbf{0.60}  \\
        \hline
        \end{tabular}
        \end{table}

    \subsection{Limitations}\label{ssec:limitations}

\begin{figure}[ht!]
        \begin{center}
        {\includegraphics[width=1.0\linewidth]{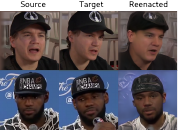}}
        \end{center}
        \caption{Cases where the reconstruction of facial accessories like hats fails. The first two columns show the source and target images, while the reenacted images are presented on the last column.}
        \label{fig:limitations}
        \end{figure}

        As shown both in our quantitative and qualitative results, our method is able to efficiently reenact the source faces, by preserving the source identity characteristics and by faithfully transferring the target head pose and expression. Our proposed method, which is based on a pre-trained StyleGAN2 model, enables both self and cross-subject reenactment using only one source frame and without any further subject fine-tuning. The proposed joint training scheme of the real image encoder $\mathcal{E}_w$ and the direction matrix $\mathbf{A}$ enables more accurate identity reconstruction and facial image editing without many visual artifacts, especially on the challenging task of extreme head poses. Additionally, the refinement of StyleGAN2's feature space $\mathcal{F}$ enables better reconstruction of various image details including background, hair style/color and facial accessories, resulting in visually more realistic images. Nevertheless, in Fig.~\ref{fig:limitations} we observe that especially on hair accessories, such as hats that are underrepresented on the training dataset, our method is not able to faithfully reconstruct every detail when editing the head pose orientation.

%%%%%%%%%%%%%%%%%%%%%%%%%%%%%%%%%%%%%%%%%%%%%%%%%%%%%%%%%%%%%%%%%%%%%%%%%%%%%%%%
%%                                                                            %%
%%                             [ Conclusions ]                                %%
%%                                                                            %%
%%%%%%%%%%%%%%%%%%%%%%%%%%%%%%%%%%%%%%%%%%%%%%%%%%%%%%%%%%%%%%%%%%%%%%%%%%%%%%%%
\section{Conclusions}
 
    In this paper, we presented a novel approach towards neural head/face reenactment using a 3D shape model to learn disentangled directions of head pose and expression in the latent GAN space. This approach comes with specific advantages, such as the use of powerful pre-trained GANs and 3D shape models, which have been thoroughly developed and studied by the research community during the past years. Our method is able to successfully disentangle the facial movements and the appearance of the input images leveraging the disentangled properties of the pre-trained StyleGAN2 model. Consequently, our framework effectively mimics the target head pose and expression without transferring identity details from the driving images. Additionally, our method features several favorable properties including one-shot face reenactment without the need for further subject-specific fine-tuning. It also allows for improved cross-subject reenactment through the proposed upaired data training with synthetic and real images. While our method demonstrates compelling results, it relies on the capabilities of StyleGAN2 model, which is bounded by the distribution of the training dataset. If the dataset lacks diversity in terms of complex backgrounds, facial accessories like hats, glasses e.t.c, this can affect our model's ability to generalize well to more complex datasets. This limitation highlights the importance of using more diverse video datasets during the training of the generative models.
    
    Finally, we acknowledge that although face reenactment can be used in a variety of applications such as art, entertainment, video conferencing etc., it can also be applied for malicious purposes, including deepfake creation, that could potentially harm individuals and the society. It is important for the researchers on our field to be aware of the potential risks and promote the responsible use of this technology.
   
\bmhead{Data Availability}
The VoxCeleb1 and VoxCeleb2 video datasets are publicly available at \url{https://www.robots.ox.ac.uk/~vgg/data/voxceleb/index.html}. One possible issue of using these two datasets is that videos might be missing or taken down from YouTube. The FaceForensics and the 300-VW datasets are available upon the acceptance of End User License forms at \url{https://github.com/ondyari/FaceForensic} and \url{https://ibug.doc.ic.ac.uk/resources/300-VW/}, respectively.

% \newpage
\bmhead{Supplementary information}
We provide an accompanying video as supplementary material that shows additional comparisons on video datasets.  
% If your article has accompanying supplementary file/s please state so here. 
% Authors reporting data from electrophoretic gels and blots should supply the full unprocessed scans for key as part of their Supplementary information. This may be requested by the editorial team/s if it is missing.

%%%%%%%%%%%%%%%%%%%%%%%%%%%%%%%%%%%%%%%%%%%%%%%%%%%%%%%%%%%%%%%%%%%%%%%%%%%%%%%%
%%                                                                            %%
%%                           [ Acknowledgments ]                              %%
%%                                                                            %%
%%%%%%%%%%%%%%%%%%%%%%%%%%%%%%%%%%%%%%%%%%%%%%%%%%%%%%%%%%%%%%%%%%%%%%%%%%%%%%%%
\bmhead{Acknowledgments}
This work has been supported by the EU H2020 AI4Media No. 951911 project.


%%%%%%%%%%%%%%%%%%%%%%%%%%%%%%%%%%%%%%%%%%%%%%%%%%%%%%%%%%%%%%%%%%%%%%%%%%%%%%%%
%%                                                                            %%
%%                              [ Appendix ]                                  %%
%%                                                                            %%
%%%%%%%%%%%%%%%%%%%%%%%%%%%%%%%%%%%%%%%%%%%%%%%%%%%%%%%%%%%%%%%%%%%%%%%%%%%%%%%%
\begin{appendices}

\section{}

    In this appendix, we first provide an analysis of the discovered directions in the latent space in App.~\ref{subsec:analysis} and we describe in detail the calculation of the shape losses in App.~\ref{subsec:shape_losses}. Additionally, we show results of our method on the task of facial attribute editing in App.~\ref{subsec:image_edit}. In App.~\ref{subsec:bechmark}, we provide details about the benchmark datasets used to evaluate our method on large head pose variations. Finally, in App.~\ref{subsec:results}, we compare the proposed framework with state-of-the-art methods for synthetic image editing on FFHQ dataset~\cite{karras2019style} and we show comparisons on real image editing against five methods that perform real image inversion using the feature space of StyleGAN2~\cite{karras2020analyzing}. Moreover, we provide additional quantitative and qualitative results both on the VoxCeleb1~\cite{Nagrani17} and the VoxCeleb2~\cite{Chung18b} datasets and we show additional results on the FaceForensics~\cite{roessler2018faceforensics} and the 300-VW~\cite{shen2015first} datasets. 

    \subsection{Analysis of the learned directions}\label{subsec:analysis}
        \subsubsection{Head pose/expression parameter vector} \label{subsubsec:facial_pose}
        
        The elements of $\mathbf{p}=[\mathbf{p}_{\theta}, \mathbf{p}_{e}]$, i.e., the head pose $\mathbf{p}_{\theta}$ and the expression $\mathbf{p}_{e}$ coefficients, are typically in different ranges of values. That is, head pose $\mathbf{p}_{\theta}$ is given in terms of the three Euler angles (i.e., yaw, pitch, and roll) in degrees (i.e., in the range $[-90,90]$), while the expression coefficients $\mathbf{p}_{e}$ are given in the range of $[-2,2]$ with the vast majority ($99\%$) of samples in VoxCeleb1 dataset being within the range of $[-1,1]$. In order to bring each element of $\mathbf{p}=[\mathbf{p}_{\theta}, \mathbf{p}_{e}]$ into a common range of values $[-a,a]$, we sampled 10,000 synthetic facial images and calculated the corresponding values for $\mathbf{p}_{\theta}$ and $\mathbf{p}_{e}$ using the pre-trained DECA~\cite{feng2020deca} network. We then re-scaled each element $x$ of $\mathbf{p}$ in $[-a,a]$ using min-max scaling; i.e., $\hat{x} = \frac{x - x_{min}}{x-x_{max}} \times 2a - a$. This way, we guarantee that each component contributes evenly to the overall facial representation, regardless of its original range, providing stability in the training process. The specific re-scaling range, i.e., $[-6,6]$, is practically imposed by the StyleGAN's latent space, as~\cite{voynov2020unsupervised} originally pointed out, meaning that traversing the latent space outside this range, often leads to severe degradation in the quality of the generated images, since latent codes lie in regions of low density.

        \subsubsection{Linearity} In this work, we discover the disentangled directions that control the head pose and the expression by optimising a matrix $\mathbf{A}$ so that:
        \begin{equation}\label{Eq:A1_Dw}
            \Delta\mathbf{w}=\mathbf{A}\Delta\mathbf{p}, 
        \end{equation}
        where $\Delta\mathbf{w}$ denotes a shift in the latent space and $\Delta\mathbf{p}$ denotes the corresponding change in the parameters space. That is, independently of the source attributes, we assume linearity between a shift $\Delta\mathbf{w}$ that is applied to an arbitrary code $\mathbf{w}$ and the induced change $\Delta\mathbf{p}$ in the parameter space -- i.e., the change between the source and the reenacted attributes.
        
\begin{figure*}
            \begin{center}
            \includegraphics[width=0.7\textwidth]{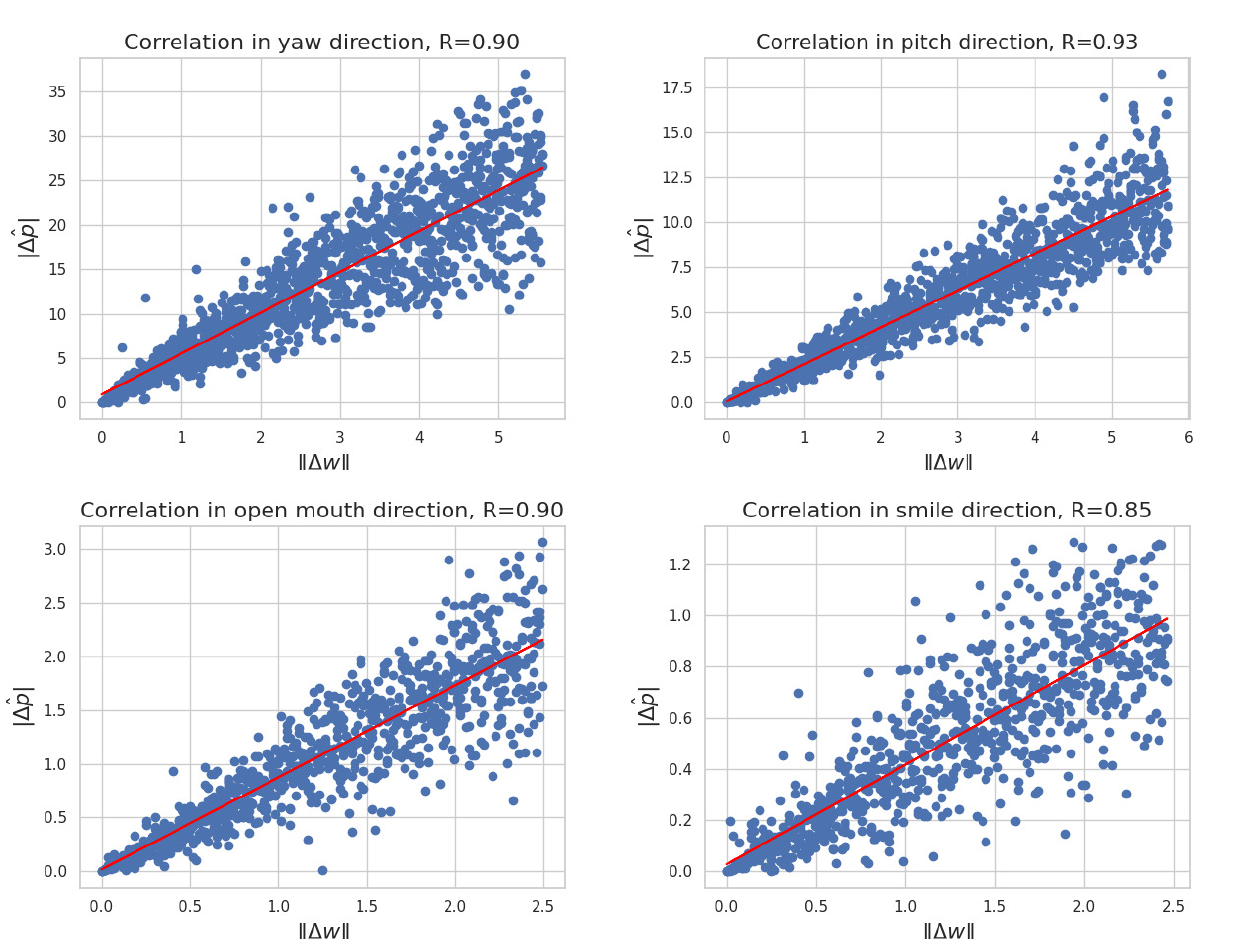}
            \end{center}
              \caption{Analysis of the correlation between shifts $\Vert\Delta\mathbf{w}\Vert$ in the latent space and the predicted changes $\lvert\hat{\Delta\mathbf{p}}\rvert$ in the parameters space. We show results of four different attributes (yaw and pitch angles, smile, and open mouth). In all attributes the correlation is high, indicating strong linear relationship.}  
            \label{fig:linearity}
        \end{figure*}

\begin{figure}[!h]
            \begin{center}
            \includegraphics[width=1.0\linewidth]{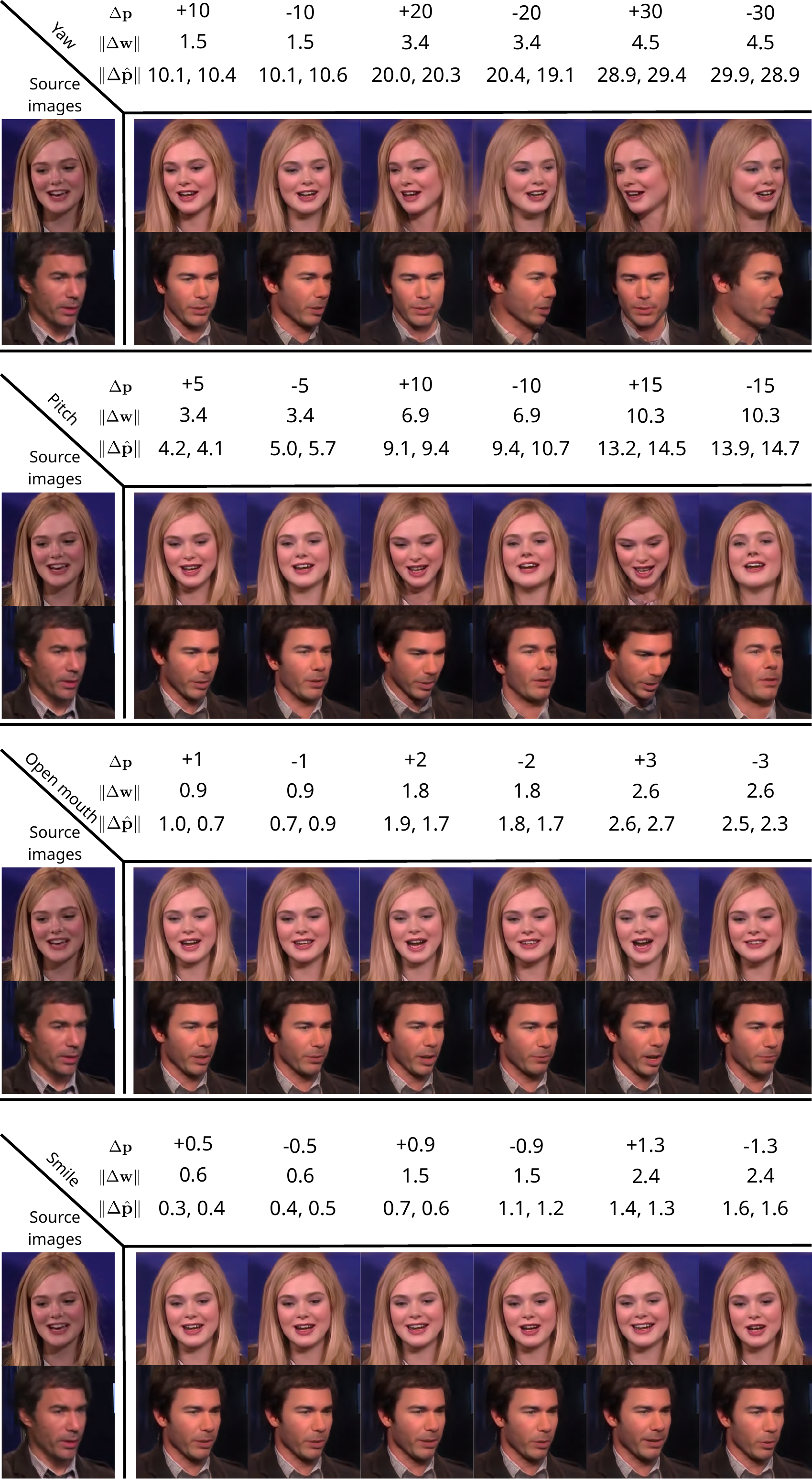}
            \end{center}
            \caption{Visual results illustrating the strongly linear relationship between $\Vert\Delta\mathbf{p}\Vert$ and $\Vert\Delta\mathbf{w}\Vert$. Specifically, given two different input images and ground truth changes $\Vert\Delta\mathbf{p}\Vert$ in the parameter space, we calculate the corresponding $\Vert\Delta\mathbf{w}\Vert$ shift in the latent space and the predicted changes $\Vert\Delta\hat{\mathbf{p}}\Vert$ between the source and shifted images. We note that a similar shift $\Vert\Delta\mathbf{w}\Vert$ corresponds to a similar change in the parameter space independently of the facial attributes of the source images.}
            \label{fig:linearity_A1}
        \end{figure}

        Several recent methods propose to learn linear directions in the latent space of StyleGAN~\cite{voynov2020unsupervised,shen2020interfacegan,shen2020closedform} in order to perform synthetic image editing, based on the fact that the $\mathcal{W}$ latent space of StyleGAN~\cite{karras2019style} has been designed to be linear and disentangled. Furthermore, \cite{nitzan2021large} provide a comprehensive analysis on the existence of linear relations between the magnitude of change in the semantic attributes (e.g., head orientation, smile, etc) and the traversal distance along the corresponding \textit{linear} latent paths. In order to further support our hypothesis (i.e., Eq.~\ref{Eq:A1_Dw}), we perform a similar analysis by examining the correlation between random shifts in the latent space, $\Delta\mathbf{w}$, and the predicted shifts in the parameters space, $\hat{\Delta \mathbf{p}}$. Specifically, given a \textit{known} change $\Delta\mathbf{p}$, we calculate the corresponding $\Delta\mathbf{w}$ using Eq.~\ref{Eq:A1_Dw} and we apply this change (i.e., $\Delta\mathbf{w}$) onto random latent codes of images with different attributes. Then, we calculate the \textit{predicted} change $\hat{\Delta\mathbf{p}}$ between the source and the reenacted images. In Fig.~\ref{fig:linearity} we demonstrate the results of our analysis in four different attributes, namely, yaw angle, pitch angle, smile, and open mouth. In all attributes, the calculated correlation is close to 0.9 indicating strong linear relationship. Finally, additional visual results of two different subjects in different head poses and expressions are depicted in Fig.~\ref{fig:linearity_A1}. Specifically, we show the ground truth change $\Vert\Delta\mathbf{p}\Vert$ in the parameter space, the corresponding $\Vert\Delta\mathbf{w}\Vert$, and the predicted changes $\Vert\Delta\hat{\mathbf{p}}\Vert$ between the source and shifted images. Above the presented images in the row where we report $\Vert\Delta\hat{\mathbf{p}}\Vert$ the two values separated by commas correspond to the subjects depicted in the first and second row. As shown, a change $\Vert\Delta\mathbf{w}\Vert$ corresponds to a similar change in the parameter space $\Vert\Delta\hat{\mathbf{p}}\Vert$ independently of the facial attributes of the source images.
        
        \subsubsection{Disentanglement}\label{subsubsec:disent} 
\begin{figure*}[t]
        \begin{subfigure}[b]{1.0\textwidth}
          \centering
          % include first image
          \includegraphics[width=1.0\linewidth]{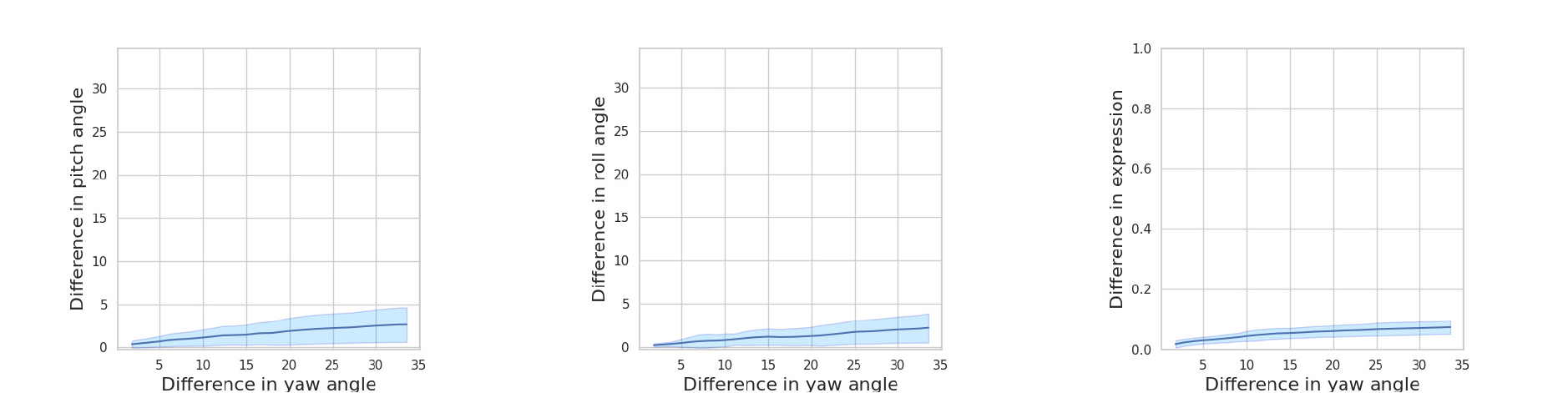}  
          \caption{L1 distance in pitch, roll angles (in degrees) and expression ($\mathbf{p}_{e}$ coefficients) when transferring only the yaw angle from the target images.}
          \label{fig:dis_yaw}
        \end{subfigure}
        \begin{subfigure}[b]{1.0\textwidth}
          \centering
          % include second image
          \includegraphics[width=1.0\textwidth]{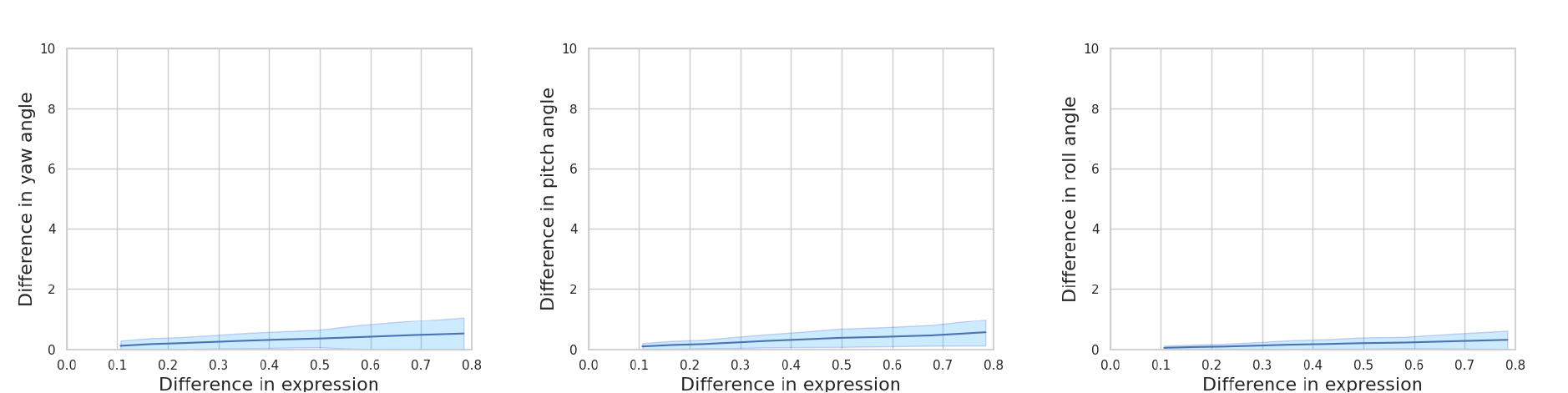}  
          \caption{L1 distance in yaw, pitch and roll angles (in degrees) when transferring only the smile expression from the target images.}
          \label{fig:dis_smile}
        \end{subfigure}
        \caption{Difference between the source and reenacted facial attributes when transferring only one facial attribute (e.g., yaw angle and smile expression) from the target images.}
        \label{fig:disentanglement}
        \end{figure*}

\begin{figure}[h]
            \begin{center}
            \includegraphics[width=1.0\linewidth]{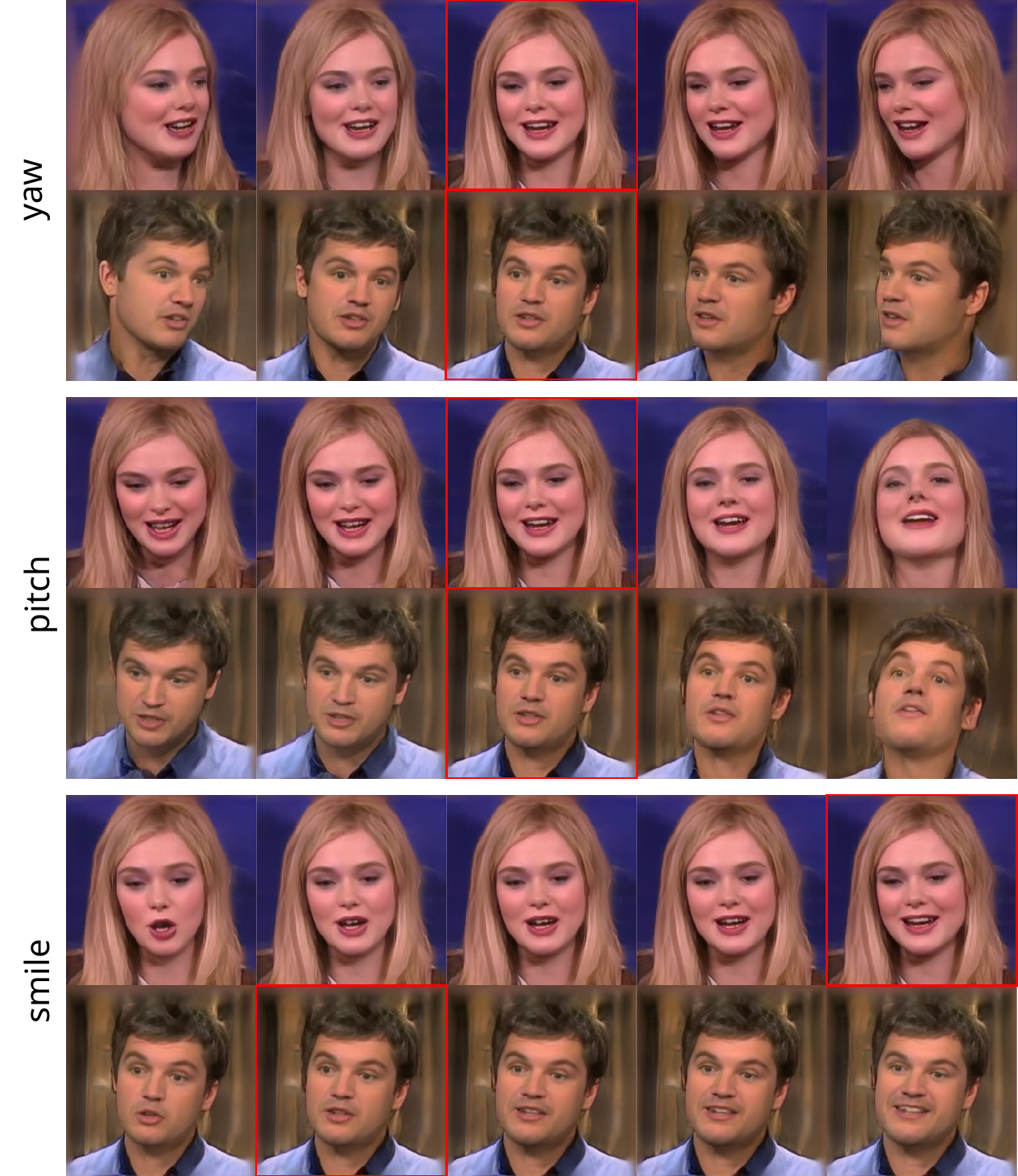}
            \end{center}
            \caption{Visual examples of editing only one facial attribute, namely yaw and pitch angles, and smile. The source images are depicted inside the red boxes.}
            \label{fig:disent_A2}
        \end{figure}

        Following the common understanding of disentanglement in the area of GANs~\cite{chen2016infogan,deng2020disentangled,karras2019style}, we refer to a disentangled latent direction when travelling across it leads to image generations where only a single attribute changes. To assess the directions learnt by our method in terms of disentanglement, in Fig.~\ref{fig:disentanglement} we illustrate the differences between the source and reenacted attributes when changing a single attribute. In Fig.~\ref{fig:dis_yaw}, we only transfer the yaw angle from the target image, while in Fig.~\ref{fig:dis_smile} we only transfer the smile expression from the target image. We observe that the differences between the rest of the attributes (i.e., pitch, roll, and expression in Fig.~\ref{fig:dis_yaw} and yaw, pitch, and roll in Fig.~\ref{fig:dis_smile}) are clearly small, which indicates that the discovered directions are disentangled. We note that these plots were calculated using 2000 random image pairs. In Fig.~\ref{fig:dis_yaw}, we show the differences in yaw angle that were calculated as the absolute difference between the source and the target yaw angles (measured in degrees), while the differences in the \textit{unchanged} attributes were calculated between the source and reenacted images. In a similar way, in Fig.~\ref{fig:dis_smile} we show the differences in expression that were calculated as the absolute difference between the source and the target expression ($\mathbf{p}_e$ coefficients). Moreover, in Fig.~\ref{fig:disent_A2} we demonstrate visual results of editing only one direction, namely yaw, pitch angles and smile. As shown, when altering the head pose, i.e., yaw and pitch angles, all other facial attributes, i.e., facial expressions, remain unchanged. Additionally, when altering the smile expression, we observe changes only around the mouth area, while head orientation remains the same. In more detail, in the first subject where smile is controlled (row 5), the eyes remain closed despite editing the smile expression, while in the second subject (row 6) the raised brows remain unaffected.
       
        Finally, in order to encourage better disentanglement between the facial attributes that we control, during training we propose to change only one attribute on $50\%$ of the training samples within each batch. To validate the effectiveness of the above training choice, in Table~\ref{table:disentaglement} we compare two models trained on synthetic images and report results indicating with ``True" the model trained with single attribute change and ``False" the model without the single attribute change. Specifically, we change only one attribute, namely the yaw, pitch, or roll head rotation angles, or one of the expression coefficients ($e_i$, $i=1,\ldots,12$). We then calculate and report the error, i.e., the $l1$-distance between the source and the reenacted attributes that should remain unchanged. For instance, when changing only the yaw angle, then both the pitch and the roll angles, as well as the expressions should remain the same as those of the source image. We note that regarding the expression error we report the mean error across all expressions. When we alter a specific expression $e_i$, we calculate the expression error by excluding that particular expression, as denoted by the last column of Table~\ref{table:disentaglement}. As shown, adopting this training strategy leads to better disentanglement with respect to all the 3 Euler angles and the 12 facial expressions.

\begin{table*}[h]
\centering
\caption{Ablation on the impact of single attribute change during training}.\label{table:disentaglement}
\begin{tabular}{|cc|c|c|c|c|c|}
\hline
\multicolumn{2}{|c|}{\multirow{2}{*}{\begin{tabular}[c]{@{}c@{}}Head pose\\ changing\end{tabular}}} & \multirow{2}{*}{\begin{tabular}[c]{@{}c@{}}Single attribute\\ change during training\end{tabular}}  & \multicolumn{4}{c|}{Absolute Error ($\downarrow$)}                                                                                                                                                                             \\ \hhline{~~~----} %\cline{4-7} 
\multicolumn{2}{|c|}{} &     & \multicolumn{1}{c|}{Yaw}   & \multicolumn{1}{c|}{Pitch}  & \multicolumn{1}{c|}{Roll}  & $\frac{1}{12}\sum_{t=1}^{12}e_i$ \\  \hline
 
\multicolumn{2}{|c|}{\multirow{2}{*}{Yaw}} & True & - &\textbf{1.2}  & \textbf{1.3} & \textbf{0.85} \\ 
\multicolumn{2}{|c|}{}                     & False & - & 1.3  & 1.4 & 0.88 \\
\hline

\multicolumn{2}{|c|}{\multirow{2}{*}{Pitch}} & True & \textbf{1.0} & - & \textbf{0.7} & \textbf{0.58}\\
\multicolumn{2}{|c|}{}  & False & 1.2 & - & 0.9 & 0.65 \\                          
\hline

\multicolumn{2}{|c|}{\multirow{2}{*}{Roll}} & True & \textbf{1.5} & \textbf{0.9} & - & \textbf{0.53}\\
\multicolumn{2}{|c|}{} & False & 1.6 & \textbf{0.9} & - &  0.57 \\
\hline

\multicolumn{2}{|c|}{\begin{tabular}[c]{@{}c@{}}Expression\\ changing\end{tabular}}                 & \multirow{2}{*}{\begin{tabular}[c]{@{}c@{}}Single expression\\ change during training\end{tabular}} & \multicolumn{1}{c|}{\multirow{2}{*}{Yaw}} & \multicolumn{1}{c|}{\multirow{2}{*}{Pitch}} & \multicolumn{1}{c|}{\multirow{2}{*}{Roll}} & \multirow{2}{*}{$\frac{1}{12}\sum_{\substack{t=0 \\  t\neq i}}^{12}e_i$} \\ \hhline{--~}%\cline{1-2}
\multicolumn{1}{|c|}{$i$}  & $e_i$  &       & \multicolumn{1}{c|}{}   & \multicolumn{1}{c|}{}                       & \multicolumn{1}{c|}{}                      &                                              \\ \hline

\multicolumn{1}{|c|}{\multirow{2}{*}{1}}  & \multirow{2}{*}{``Expression\_1''} & True  & \textbf{0.3} & \textbf{0.2} & \textbf{0.2} & \textbf{0.49} \\
\multicolumn{1}{|c|}{} &  & False & 0.7 & 0.5 & 0.4 & 0.53 \\
\hline

\multicolumn{1}{|c|}{\multirow{2}{*}{2}}  & \multirow{2}{*}{``Expression\_2''} & True  & \textbf{0.4} & \textbf{0.2} & \textbf{0.1} & \textbf{0.27} \\
\multicolumn{1}{|c|}{} &  & False & 0.8 & 0.5 & 0.3 & 0.40 \\
\hline

\multicolumn{1}{|c|}{\multirow{2}{*}{3}}  & \multirow{2}{*}{``Expression\_3''} & True  & \textbf{0.3} & \textbf{0.2} & \textbf{0.1} & \textbf{0.18} \\
\multicolumn{1}{|c|}{} &  & False & 0.7 & 0.4 & 0.3 & 0.30 \\
\hline

\multicolumn{1}{|c|}{\multirow{2}{*}{4}}  & \multirow{2}{*}{``Expression\_4''} & True  & \textbf{0.4} & \textbf{0.2} & \textbf{0.2} & \textbf{0.16} \\
\multicolumn{1}{|c|}{} &  & False & 0.7 & 0.4 & 0.5 & 0.30 \\
\hline

\multicolumn{1}{|c|}{\multirow{2}{*}{5}}  & \multirow{2}{*}{``Expression\_5''} & True  & \textbf{0.3} & \textbf{0.2} & \textbf{0.1} & \textbf{0.25} \\
\multicolumn{1}{|c|}{} &  & False & 1.0 & 0.4 & 0.5 & 0.50 \\
\hline

\multicolumn{1}{|c|}{\multirow{2}{*}{6}}  & \multirow{2}{*}{``Expression\_6''} & True  & \textbf{0.4} & \textbf{0.2} & \textbf{0.1} & \textbf{0.35}\\
\multicolumn{1}{|c|}{} &  & False & 0.7 & 1.0 & 0.4 & 0.50 \\
\hline

\multicolumn{1}{|c|}{\multirow{2}{*}{7}}  & \multirow{2}{*}{``Expression\_7''} & True  & \textbf{0.4} & \textbf{0.2} & \textbf{0.1} & \textbf{0.26} \\
\multicolumn{1}{|c|}{} &  & False & 0.7 & 1.0 & 0.4 & 0.36 \\
\hline

\multicolumn{1}{|c|}{\multirow{2}{*}{8}}  & \multirow{2}{*}{``Expression\_8''} & True  & \textbf{0.3} & \textbf{0.2} & \textbf{0.1} & \textbf{0.11} \\
\multicolumn{1}{|c|}{} &  & False & 0.7 & 1.0 & 0.4 & 0.25 \\
\hline

\multicolumn{1}{|c|}{\multirow{2}{*}{9}}  & \multirow{2}{*}{``Expression\_9''} & True  & \textbf{0.3} & \textbf{0.2} & \textbf{0.1} & \textbf{0.23} \\
\multicolumn{1}{|c|}{} &  & False & 0.7 & 1.0 & 0.4 & 0.37  \\
\hline

\multicolumn{1}{|c|}{\multirow{2}{*}{10}}  & \multirow{2}{*}{``Expression\_10''} & True  & \textbf{0.4} & \textbf{0.2} & \textbf{0.1} & \textbf{0.18} \\
\multicolumn{1}{|c|}{} &  & False & 0.7 & 1.0 & 0.4 & 0.30 \\
\hline

\multicolumn{1}{|c|}{\multirow{2}{*}{11}}  & \multirow{2}{*}{``Expression\_11''} & True  & \textbf{0.3} & \textbf{0.2} & \textbf{0.1} & \textbf{0.18}\\
\multicolumn{1}{|c|}{} &  & False & 0.7 & 1.0 & 0.4 & 0.30 \\
\hline

\multicolumn{1}{|c|}{\multirow{2}{*}{12}}  & \multirow{2}{*}{``Expression\_12''} & True  & \textbf{0.4} & \textbf{0.2} & \textbf{0.2} & \textbf{0.16}\\
\multicolumn{1}{|c|}{} &  & False & 0.7 & 1.0 & 0.4 & 0.35 \\
\hline

\end{tabular}
\end{table*}

    \subsection{Shape losses}\label{subsec:shape_losses}

\begin{figure}
        \begin{center}
        {\includegraphics[width=0.49\textwidth]{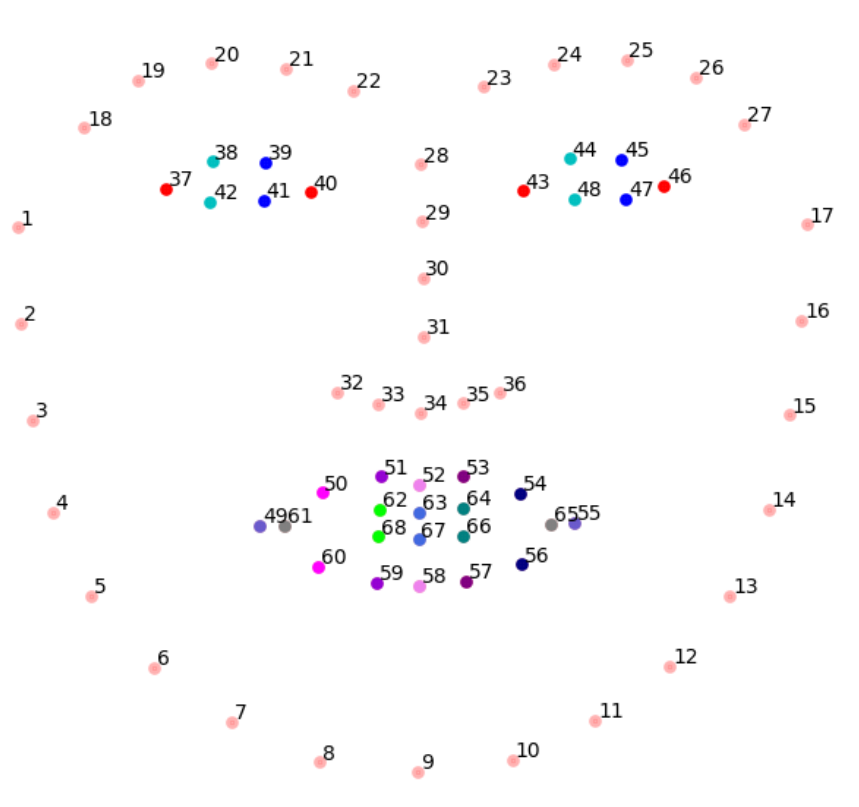}}
          \caption{Depiction of the landmark pairs $P_{eye}$ and $P_{mouth}$ that contribute to the corresponding losses $\mathcal{L}_{eye}$ and $\mathcal{L}_{mouth}$. The landmarks of each pair are drawn with the same color.} 
        \label{fig:landmarks_pairs}
        \end{center}
        \end{figure}

        In order to transfer the target head pose and expression to the source face, we calculate the \textit{reenactment loss} as:
        \begin{equation}\label{eq:loss_cat}
        \mathcal{L}_{r} =  \mathcal{L}_{sh} + \mathcal{L}_{eye} + \mathcal{L}_{mouth}, 
        \end{equation}
        where $\mathcal{L}_{sh} $ is the shape loss and $\mathcal{L}_{eye}$, $\mathcal{L}_{mouth}$ the eye and mouth loss, respectively. As shown in our ablation studies (Sect.~\ref{ssec:ablation}), $\mathcal{L}_{eye}$, $\mathcal{L}_{mouth}$ losses contribute to more accurate expression transfer from the target face to the source face. Specifically, eye loss compares the inner distances $\mathrm{d} = \|(\cdot, \cdot)\|_1$ of the eye landmark pairs (defined as $P_{eye}$) of upper and lower eyelids between the reenacted ($\mathbf{S}_r$) and reconstructed ground-truth ($\mathbf{S}_{gt}$) shapes:
        \begin{equation}\label{eq:eye_loss}
        \mathcal{L}_{eye} = \sum_{(i,j)\in P_{eye}} \left \| \mathrm{d}\big(\mathbf{S}_r(i),\mathbf{S}_r(j)\big) - \mathrm{d}\big(\mathbf{S}_{gt}(i),\mathbf{S}_{gt}(j) \right \|,
        \end{equation}
        Similarly, mouth loss is computed between the mouth landmark pairs. In Fig.~\ref{fig:landmarks_pairs}, we show the landmark pairs that are used to calculate these losses. In more detail, $P_{eye}$ and $P_{mouth}$ are defined as:
        \begin{equation*}
        \begin{split}
        P_{eye} = \big[(37,40), (38,42), (39, 41),  \\
         (43,46), (44,48), (45,47) \big],
        \end{split}
        \end{equation*} 
        \begin{equation*}
        \begin{split}
        P_{mouth} = \big[(49,55), (50,60), (51,59), (52,58),  \\
         (53,57), (54,56), \\ 
         (61,65), (62,68), (63,67), (64,66) \big]
        \end{split}
        \end{equation*} 
        
    \subsection{Image editing}\label{subsec:image_edit}

\begin{figure}[t]
        \begin{center}
        {\includegraphics[width=1.0\linewidth]{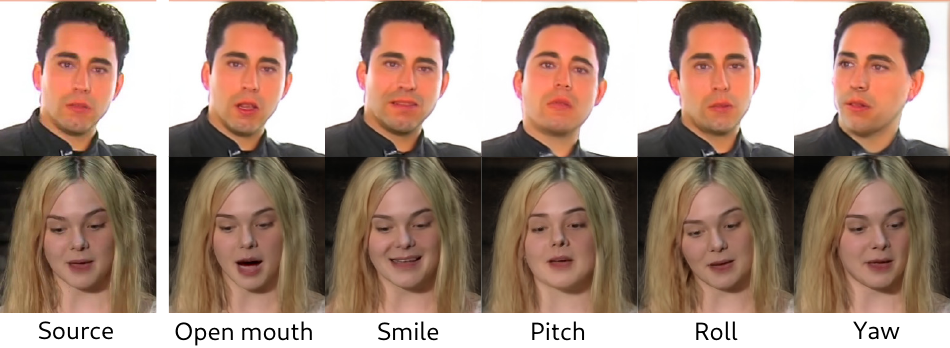}}
          \caption{Our method can perform head pose and expression editing on real images. Specifically, we are able to edit an attribute by keeping all other attributes unchanged. The first column shows the source images, while the rest columns show editings of different expressions and head poses.} 
        \label{fig:editing}
        \end{center}
        \end{figure}

\begin{figure}[h]
        \begin{center}
        {\includegraphics[width=1.0\linewidth]{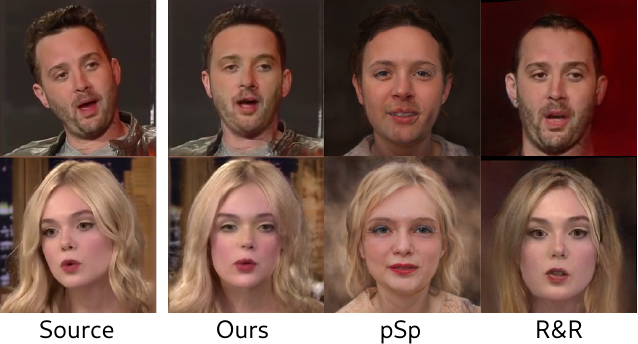}}
        \end{center}
          \caption{Face frontalization examples. We perform comparisons with pSp~\cite{richardson2021encoding} and R\&R~\cite{zhou2020rotate} and we show that our method successfully perform face frontalization by preserving the identity of the source face.}\label{fig:frontal_fig}
        \end{figure}

        Our method is able to discover the disentangled directions of head pose and expression in the latent space of StyleGAN2. Consequently, except from face reenactment, our model can perform head pose and expression editing by simply setting the desired head pose or expression. Fig.~\ref{fig:editing} illustrates results of per attribute editing. As shown, our model can alter the head pose (i.e., yaw, pitch, and roll) or the expression (e.g., open mouth, smile) by maintaining all other attributes unchanged. Similarly, our method can be used in the frontalization task. We compare our model with the methods of pSp~\cite{richardson2021encoding} and R\&R~\cite{zhou2020rotate} and we report both qualitative (Fig.~\ref{fig:frontal_fig}) and quantitative (Table~\ref{table:frontal}) results. Specifically, we randomly select 250 frames of different identities from the VoxCeleb test set and we perform frontalization. In Table~\ref{table:frontal}, we evaluate the identity preservation (CSIM) and the Average Rotation Distance (ARD) between the source and the frontalized images.
        
        \begin{table}[t]
        \caption{Quantitative results on frontalization task. We compare our method with pSp~\cite{richardson2021encoding} and R\&R~\cite{zhou2020rotate} by evaluating the identity preservation (CSIM) and the Average Rotation Distance (ARD) between the source and the frontalized images.}\label{table:frontal}
        \begin{tabular}{|c|c|c|}
            \hline
            Method & CSIM & ARD \\
            \hline
            pSp~\cite{richardson2021encoding} & 0.40 & 3.0 \\
            R\&R~\cite{zhou2020rotate} & 0.45 & 3.5\\
            Ours & \textbf{0.60} & \textbf{1.2} \\
            \hline
        \end{tabular}
        \end{table}
        
    \subsection{Benchmark datasets with large head pose variations}\label{subsec:bechmark}

\begin{figure}[h]
        \begin{center}
        \includegraphics[width=1.0\linewidth]{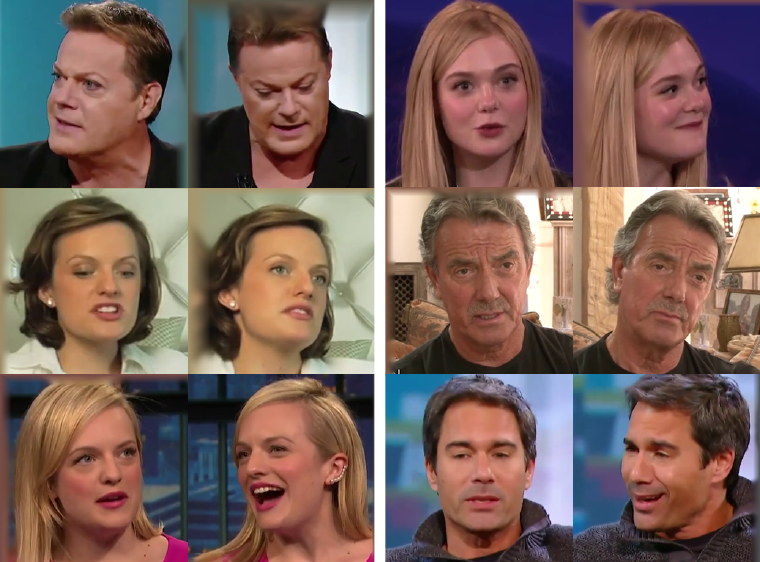}
        \end{center}
        \caption{Indicative examples of source-target image pairs from our benchmark set (Benchmark-L), where the average head pose distance is larger than $10^{\circ}$.}
        \label{fig:examples_benchmark}
    \end{figure}

\begin{figure*}[h]
        \begin{center}
        \includegraphics[width=0.8\textwidth]{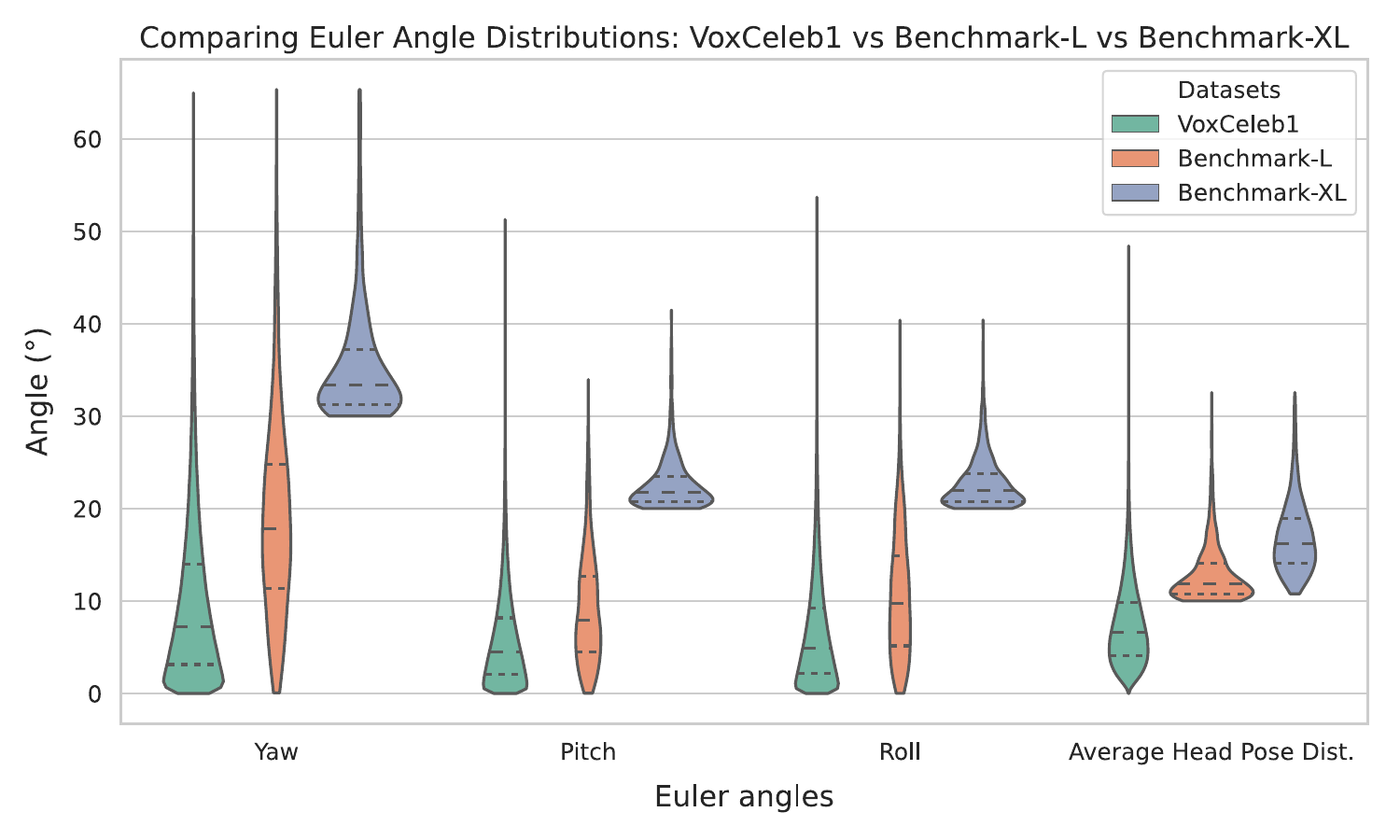}
        \end{center}
        \caption{Comparison of the distributions of the three Euler angles (yaw, pitch and roll) and the average head pose distance between the VoxCeleb1 dataset, the benchmark set, called here Benchmark-L (average head pose distance larger than $10^{\circ}$), and the new benchmark set, called here Benchmark-XL (yaw larger than $30^{\circ}$, pitch/roll larger than $20^{\circ}$)}
        \label{fig:benchmark_dist}
    \end{figure*}

    As mentioned in Sect.~\ref{subsubsec:quanti_comp}, the benchmark used in Table~\ref{table:self_reenactment_large_metrics} (Benchmark-L) in order to evaluate our method on large head pose reenactment contains 1,000 image pairs from the VoxCeleb1 dataset with head pose distance larger than $10^{\circ}$, calculated as the average $L1$ distance of the three Euler angles (yaw, pitch, roll). In Fig.~\ref{fig:benchmark_dist}, we present a comparison of the distributions of the three Euler angles (yaw, pitch and roll) and the average head pose distance between the VoxCeleb1 dataset and the aforementioned benchmark dataset (Benchmark-L). As shown Benchmark-L comprises image pairs that have larger head pose distances compared to the average head pose distance observed in the VoxCeleb1 dataset. Additionally, Fig.~\ref{fig:examples_benchmark}, illustrates some indicative example image pairs from the benchmark dataset, where there is a wide range on the head pose variations across all three Euler angles.

\begin{figure*}[h]
        \begin{center}
        \includegraphics[width=1.0\textwidth] {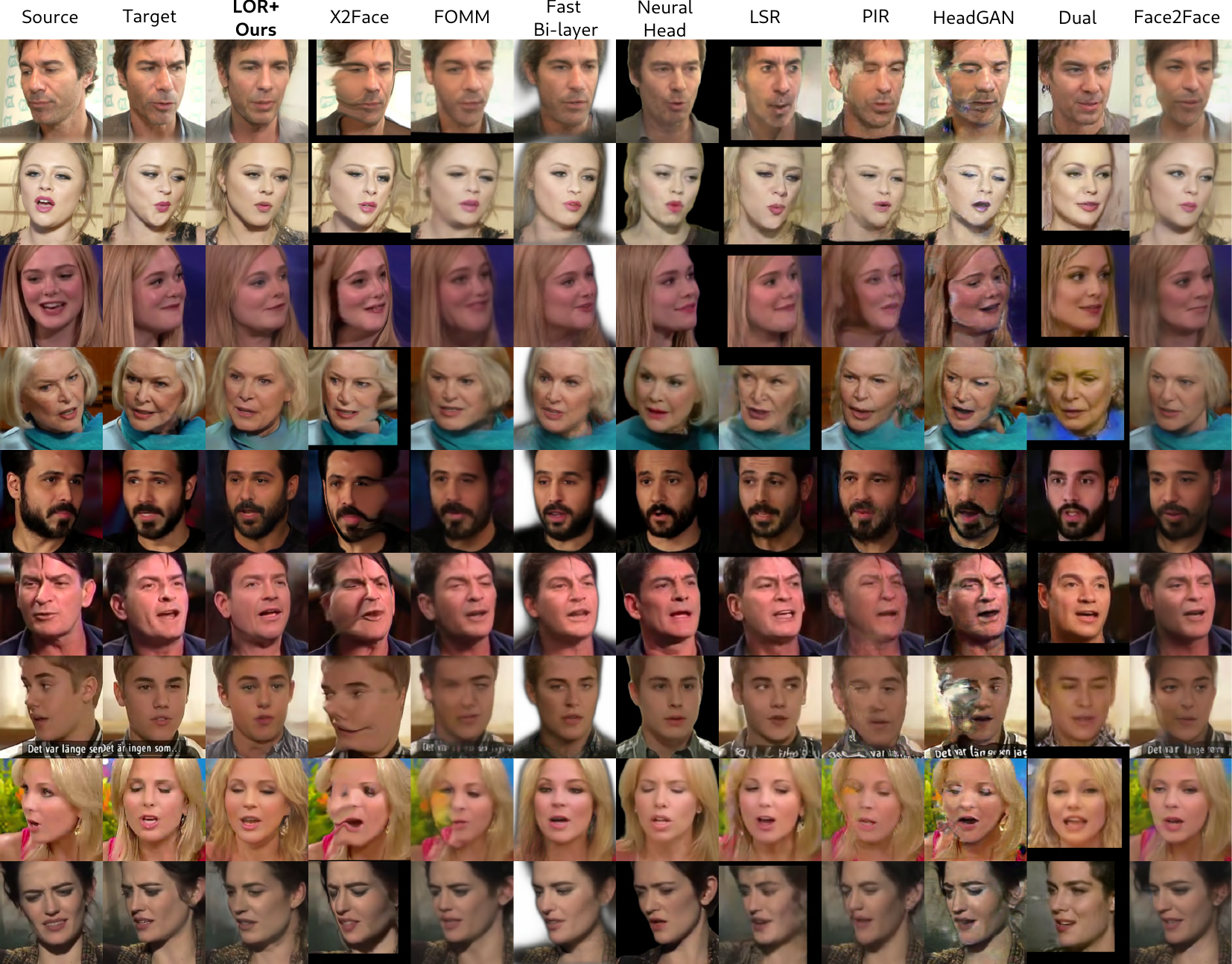}
        \end{center}
        \caption{Qualitative comparisons on the Benchmark-XL with image pairs from VoxCeleb1 and VoxCeleb2 datasets, where the distance on the yaw angle is larger than $30^{\circ}$ and on the pitch or roll angles larger than $20^{\circ}$.}
        \label{fig:comparisons_large}
    \end{figure*}

    To further validate our method on larger head pose differences we generate a new benchmark dataset (Benchmark-XL) using images from both the VoxCeleb1 and the VoxCeleb2 datasets. Specifically, we randomly select $1,000$ image pairs where the distance on the yaw angle is larger than $30^{\circ}$ and on the pitch or roll angles larger than $20^{\circ}$. As shown in Fig.~\ref{fig:benchmark_dist}, Benchmark-XL consists of image pairs with ``extreme" head pose differences compared to the distribution of the overall dataset. In Tables~\ref{table:large_pose_vox1},~\ref{table:large_pose_vox2} and in Fig.~\ref{fig:comparisons_large}, we demonstrate the quantitative and qualitative comparisons on Benchmark-XL both on VoxCeleb1 and VoxCeleb2, respectively. As shown our method is able to better preserve the identity of the source faces, successfully transfer the target head pose and expression and generate realistic images without many visual artifacts compared to the other state-of-the-art methods.

    \begin{table*}[t]
    \begin{center}
    \caption{Quantitative comparisons on the Benchmark-XL with image pairs from VoxCeleb1 dataset, where the distance on the yaw angle is larger than $30^{\circ}$ and on the pitch or roll angles larger than $20^{\circ}$.}\label{table:large_pose_vox1}
        \begin{tabular}{|c|c|c|c|c|c|c|}
        \hline
        Method & CSIM & LPIPS & FID & ARD & AED & AU-H  \\
         \hline
        X2Face~\cite{wiles2018x2face} & 0.55 & \underline{0.13} & 91.2 & 2.2 & 1.1 & 0.27\\
        FOMM~\cite{siarohin2019first} & \underline{0.56} & 0.14 & 92.6 & 2.7 & 1.2 & 0.27\\
        Fast Bi-layer~\cite{zakharov2020fast} & 0.53 & 0.19 & 113.5 & 1.4 & \underline{0.8} & \underline{0.23} \\
        Neural-Head~\cite{burkov2020neural} & 0.40 & 0.17 & 109.5 & 2.0 & \underline{0.8} & 0.25 \\
        LSR~\cite{meshry2021learned} & 0.53 & \textbf{0.12} & 78.1 & \underline{1.2} & \underline{0.8} & \underline{0.23} \\
        PIR~\cite{ren2021pirenderer} & 0.53 & 0.14 & 95.5 & 3.0 & 1.1 & 0.27 \\
        HeadGAN~\cite{doukas2020headgan} & 0.30 & 0.27  & 92.0 & 3.7  & 1.4 & 0.30  \\
        Dual~\cite{hsu2022dual} & 0.25 & 0.20 & 101.3 & 4.6 & 1.2 & 0.28 \\
        Face2Face~\cite{yang2022face2face} & 0.38 & 0.28 & \textbf{60.0} & 3.4 & 1.1 & 0.27 \\
        LOR (Ours) & 0.42  & 0.15  & 72.0 & 2.0  & 1.0  & 0.24\\
        LOR+ (Ours) &  \textbf{0.57}& \underline{0.13} & \underline{65.6} & \textbf{0.9} & \textbf{0.6} & \textbf{0.22} \\
        \hline
        \end{tabular}
    \end{center}
    \end{table*}

    \begin{table*}[t]
    \begin{center}
    \caption{Quantitative comparisons on the Benchmark-XL with image pairs from VoxCeleb2 dataset, where the distance on the yaw angle is larger than $30^{\circ}$ and on the pitch or roll angles larger than $20^{\circ}$.}\label{table:large_pose_vox2}
        \begin{tabular}{|c|c|c|c|c|c|c|}
        \hline
        Method & CSIM & LPIPS & FID & ARD & AED & AU-H  \\
         \hline
        X2Face~\cite{wiles2018x2face} & 0.45 &  0.20 & 161.4 & 8.6 & 1.4 & 0.31 \\
        FOMM~\cite{siarohin2019first} & 0.49 & \underline{0.18}  & 175.3 & 6.2 & 1.2 & 0.28\\
        Fast Bi-layer~\cite{zakharov2020fast} & 0.47 & 0.22 & 172.2 & 1.7 & 0.9 & 0.27 \\
        Neural-Head~\cite{burkov2020neural} & 0.36 & \underline{0.18} & 160.1 & 1.9 & 1.0 & \underline{0.25}  \\
        LSR~\cite{meshry2021learned} & \underline{0.51} & \textbf{0.15} & 146.6 & \underline{1.4} & \underline{0.8} & \textbf{0.24} \\
        PIR~\cite{ren2021pirenderer} & 0.42 & 0.19 & 173.1 & 4.5 & 1.2 & 0.27 \\
        HeadGAN~\cite{doukas2020headgan} & 0.28  & 0.32 & 170.2 & 2.5 & 1.4 & 0.33  \\
        Dual~\cite{hsu2022dual} & 0.22 & 0.30 & \underline{146.3} & 4.8 & 1.3 & 0.27 \\
        Face2Face~\cite{yang2022face2face} & 0.29 & 0.30 & 151.2 & 2.5 & 1.1 & 0.29 \\
        LOR (Ours) & 0.39  & 0.17  & 155.5 & 2.5  & 1.0  & 0.29 \\
        LOR+ (Ours) &  \textbf{0.53}& \underline{0.18} & \textbf{136.0} & \textbf{1.3} & \textbf{0.7} & \textbf{0.24} \\
        \hline
        \end{tabular}
    \end{center}
    \end{table*}

    \subsection{Additional results}\label{subsec:results}

        \subsubsection{Comparisons with synthetic image editing methods}
\begin{figure}[t]
        \begin{center}
        {\includegraphics[width=1.0\linewidth]{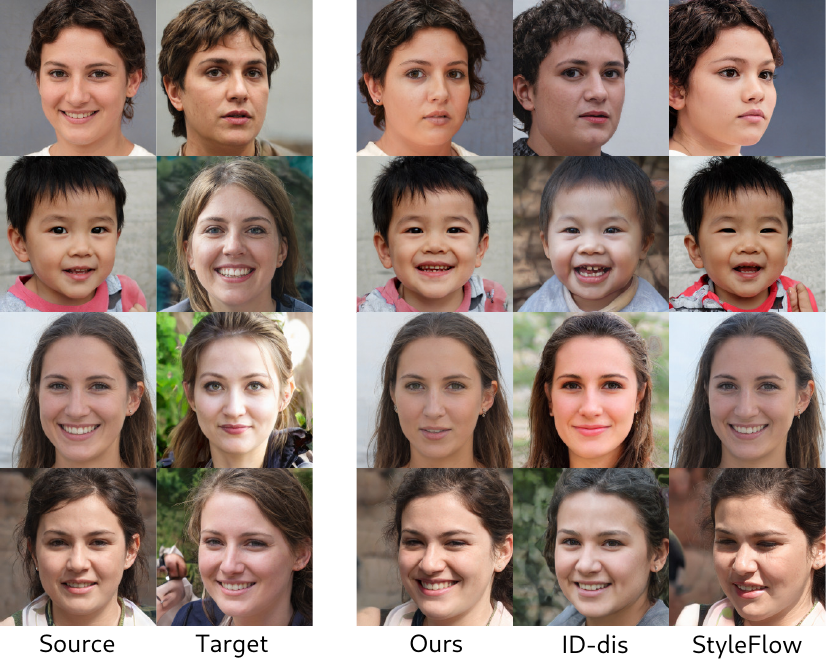}}
        \end{center}
          \caption{Qualitative comparisons against ID-disentanglement (ID-dis)~\cite{nitzan2020face} and StyleFlow~\cite{abdal2021styleflow} using random source-target pairs from the small test set provided by the authors of StyleFlow~\cite{abdal2021styleflow}.}\label{fig:comparisons_synthetic}
        \end{figure}

        In order to show the superiority of our method against methods for synthetic image editing, we compare against two state-of-the-art methods, namely ID-disentanglement~\cite{nitzan2020face} and StyleFlow~\cite{abdal2021styleflow}. The authors of ID-disentanglement~\cite{nitzan2020face} introduce a method that learns to disentangle the head pose/expression and the identity characteristics using a pre-trained StyleGAN2 on FFHQ dataset. Additionally, StyleFlow~\cite{abdal2021styleflow} is a state-of-the-art method that finds meaningful non-linear directions in the latent space of StyleGAN2 using supervision from multiple attribute classifiers and regressors. Both ID-disentanglement~\cite{nitzan2020face} and StyleFlow~\cite{abdal2021styleflow} provide pre-trained models using the StyleGAN2 generator trained on FFHQ dataset~\cite{karras2019style}. Consequently, in order to fairly compare against these methods, we train our model using synthetically generated images from StyleGAN2 generator trained on FFHQ. We compare against ID-disentanglement~\cite{nitzan2020face} and StyleFlow~\cite{abdal2021styleflow} on cross-subject reenactment using synthetic images. Specifically, we use the small test set (1000 images) provided by the authors of StyleFlow~\cite{abdal2021styleflow} and we randomly select 500 image pairs (source and target faces) to perform face reenactment. In Table~\ref{table:comparisons_synthetic_table} and in Fig.~\ref{fig:comparisons_synthetic}, we show quantitative and qualitative results of our method against ID-disentanglement~\cite{nitzan2020face} and StyleFlow~\cite{abdal2021styleflow}. As shown in Table~\ref{table:comparisons_synthetic_table} our method outperforms all other method both on identity preservation (CSIM) and on head pose/expression transfer metrics, namely ARD, AED and NME. Additionally, as illustrated in Fig.~\ref{fig:comparisons_synthetic}, our method can successfully edit the source image given the target head pose/expression, without altering the source identity. On the contrary, ID-disentanglement (ID-dis) method~\cite{nitzan2020face} is not able to preserve the source identity, while StyleFlow~\cite{abdal2021styleflow} fails to faithfully transfer the target head pose and expression. 
        %%%%%%%%%%%%%%%%%%%%%%%%%%%%%%%%%%%%%%%%%%%%%%%%%%%%%%%%%%%%
        
        \begin{table*}[t]
        \begin{center}
        \caption{Quantitative comparisons against two state-of-the-art methods for synthetic image editing, namely ID-dis~\cite{nitzan2020face} and StyleFlow~\cite{abdal2021styleflow}. For CSIM metric, higher is better ($\uparrow$), while in all other metrics lower is better ($\downarrow$).}\label{table:comparisons_synthetic_table}
        \begin{tabular}{|c|c|c|c|c|}
            \hline
            Method & CSIM   & ARD  & AED  & NME \\
            \hline
            ID-dis~\cite{nitzan2020face}  & 0.56 & 2.0 & 0.12 & 12.0\\
            StyleFlow~\cite{abdal2021styleflow} & 0.67 & 2.6 & 0.13 & 16.0 \\
            Ours  & \textbf{0.80} & \textbf{1.1} & \textbf{0.09} & \textbf{10.1} \\
            \hline
        \end{tabular}
        \end{center}
        \end{table*}
        
        \subsubsection{Comparisons with real image inversion methods}
\begin{figure*}[h]
            \begin{center}
            \includegraphics[width=0.8\textwidth]{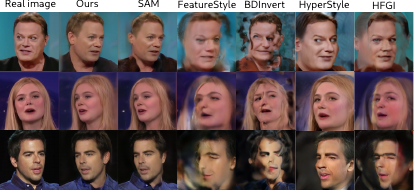}
            \end{center}
            \caption{Qualitative comparison of the proposed framework (Ours) against SAM~\cite{parmar2022spatially}, FeatureStyle ~\cite{xuyao2022}, BDInvert~\cite{kang2021gan}, HyperStyle~\cite{alaluf2022hyperstyle}, and HFGI~\cite{wang2022high} on the task of real image head pose editing.}
            \label{fig:comparisons_inversion}
        \end{figure*}

        Additionally, in order to validate that our proposed Feature Transformation module is necessary to perform image editing without producing visual artifacts when altering the feature space of StyleGAN2, we compare our method against four methods that perform real image inversion using the feature space and one method that learns to alter the weights of the StyleGAN2 generator. Specifically, we compare against SAM~\cite{parmar2022spatially}, FeatureStyle~\cite{yao2022feature}, BDInvert~\cite{kang2021gan}, HFGI~\cite{wang2022high} and HyperStyle~\cite{alaluf2022hyperstyle}. Both SAM~\cite{parmar2022spatially} and BDInvert~\cite{kang2021gan} are optimization-based approaches that refine the feature space of StyleGAN2 to perform real image inversion with better reconstruction quality. Additionally, FeatureStyle~\cite{yao2022feature} is an encoder-based method that simultaneously predicts the inverted latent code $\mathbf{w}$ and feature map $F_K$ at K\textsuperscript{th} convolution layer of StyleGAN2. Similarly, HFGI~\cite{wang2022high} predicts both the latent code $\mathbf{w}$ and the spatial feature of StyleGAN2 generator to improve the inversion quality. Finally, HyperStyle~\cite{alaluf2022hyperstyle} proposes to alter the generator's weights using a hypernetwork. In Fig.~\ref{fig:comparisons_inversion}, we demonstrate results of editing the head pose using our direction matrix $\mathbf{A}$ by first inverting the real images using the above methods. As shown, our method is the only one without visual artifacts when editing the head pose orientation. All the aforementioned methods are able to faithfully reconstruct the real images but fail on editing. 
        
        \subsubsection{Additional comparisons}
        In Table~\ref{table:user_study}, we report the results of our user study. Specifically, we ask 30 users to select the method that best reenacts the source frame on self and cross-subject reenactment tasks. For the purposes of the user study we utilise only our final model (LOR+) and we compare against X2Face~\cite{wiles2018x2face}, FOMM~\cite{siarohin2019first}, Fast bi-layer~\cite{zakharov2020fast}, Neural-Head~\cite{burkov2020neural}, LSR~\cite{meshry2021learned} and PIR~\cite{ren2021pirenderer}. As shown our method is the most preferable, by a large margin -- 52.1\% versus the 19.2\% of the second best method. 
        
\begin{figure*}
\begin{center}
{\includegraphics[width=1.0\linewidth]{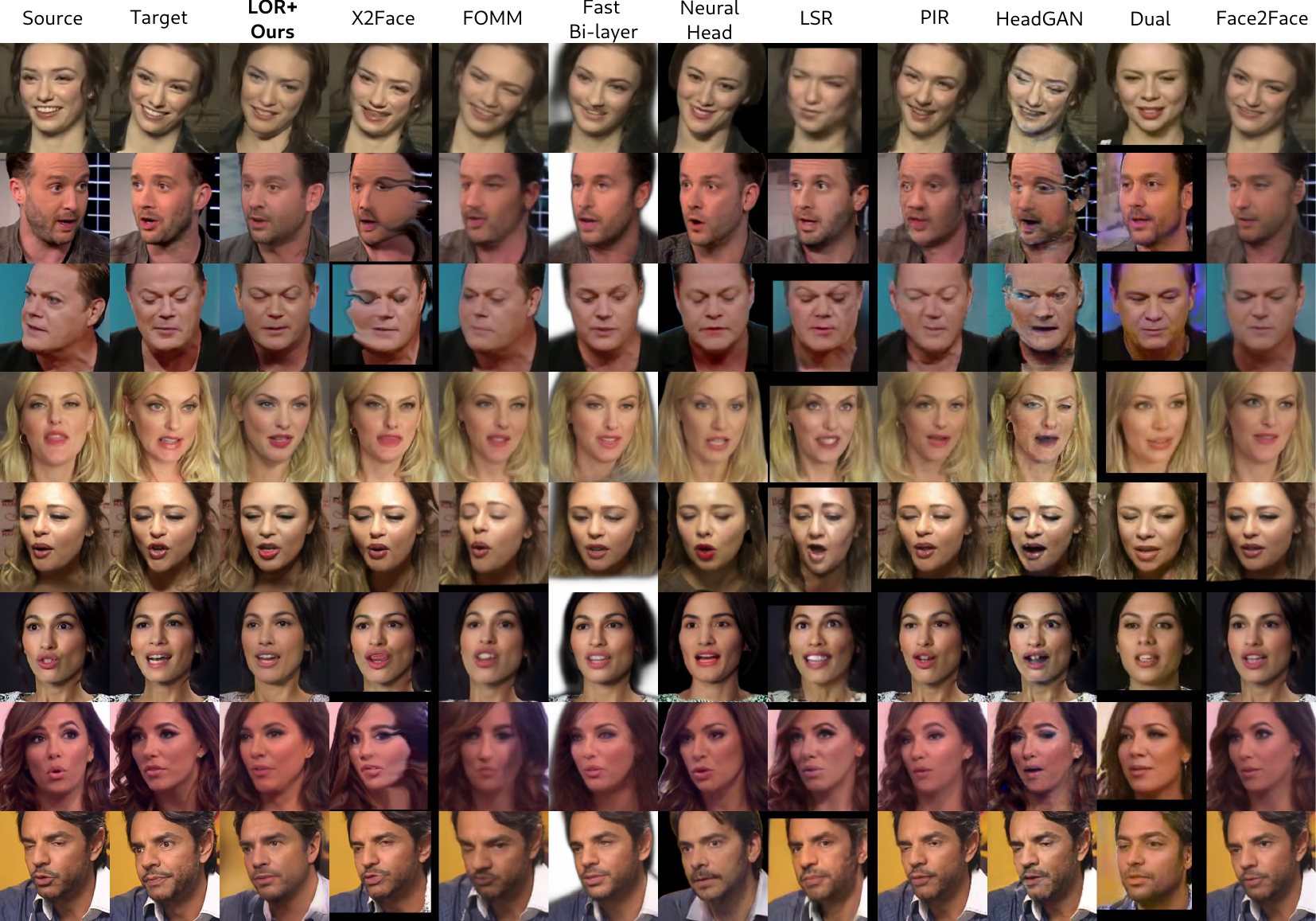}}
\end{center}
  \caption{Qualitative results and comparisons for self-reenactment on VoxCeleb1~\cite{Nagrani17} dataset. The first and second columns show the source and target faces. We compare our method against X2Face~\cite{wiles2018x2face}, FOMM~\cite{siarohin2019first}, Fast Bi-layer~\cite{zakharov2020fast}, Neural-Head~\cite{burkov2020neural}, LSR~\cite{meshry2021learned}, PIR~\cite{ren2021pirenderer}, HeadGAN~\cite{doukas2020headgan}, Dual~\cite{hsu2022dual} and Face2Face~\cite{yang2022face2face}.}\label{fig:self_vox1}
\end{figure*}

\begin{figure*}
\begin{center}
{\includegraphics[width=1.0\linewidth]{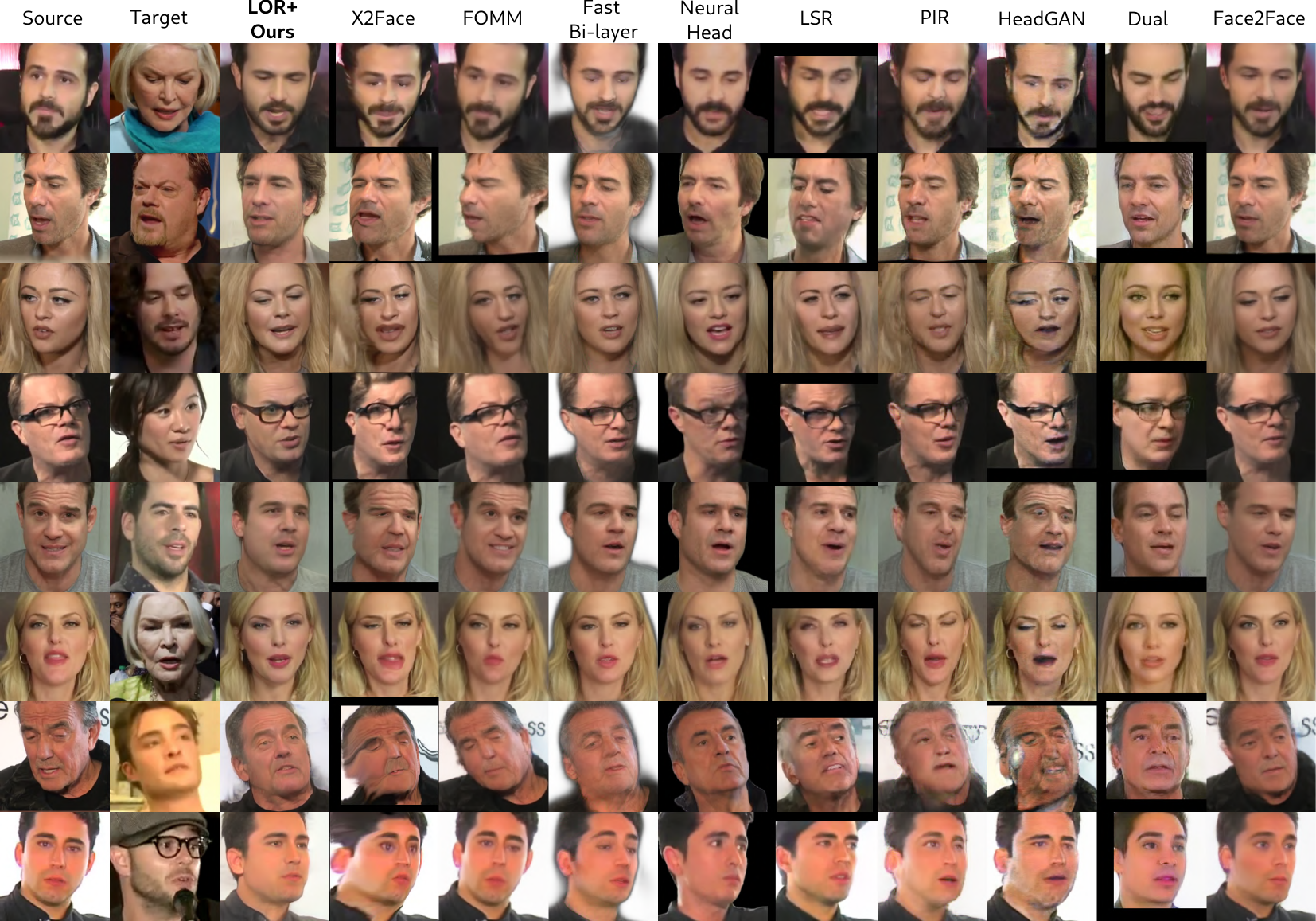}}
\end{center}
  \caption{Qualitative results and comparisons for cross-subject reenactment on VoxCeleb1~\cite{Nagrani17} dataset. The first and second columns show the source and target faces. We compare our method against X2Face~\cite{wiles2018x2face}, FOMM~\cite{siarohin2019first}, Fast Bi-layer~\cite{zakharov2020fast}, Neural-Head~\cite{burkov2020neural}, LSR~\cite{meshry2021learned}, PIR~\cite{ren2021pirenderer}, HeadGAN~\cite{doukas2020headgan}, Dual~\cite{hsu2022dual} and Face2Face~\cite{yang2022face2face}.}\label{fig:cross_vox1}
\end{figure*}

\begin{figure*}
\begin{center}
{\includegraphics[width=1.0\linewidth]{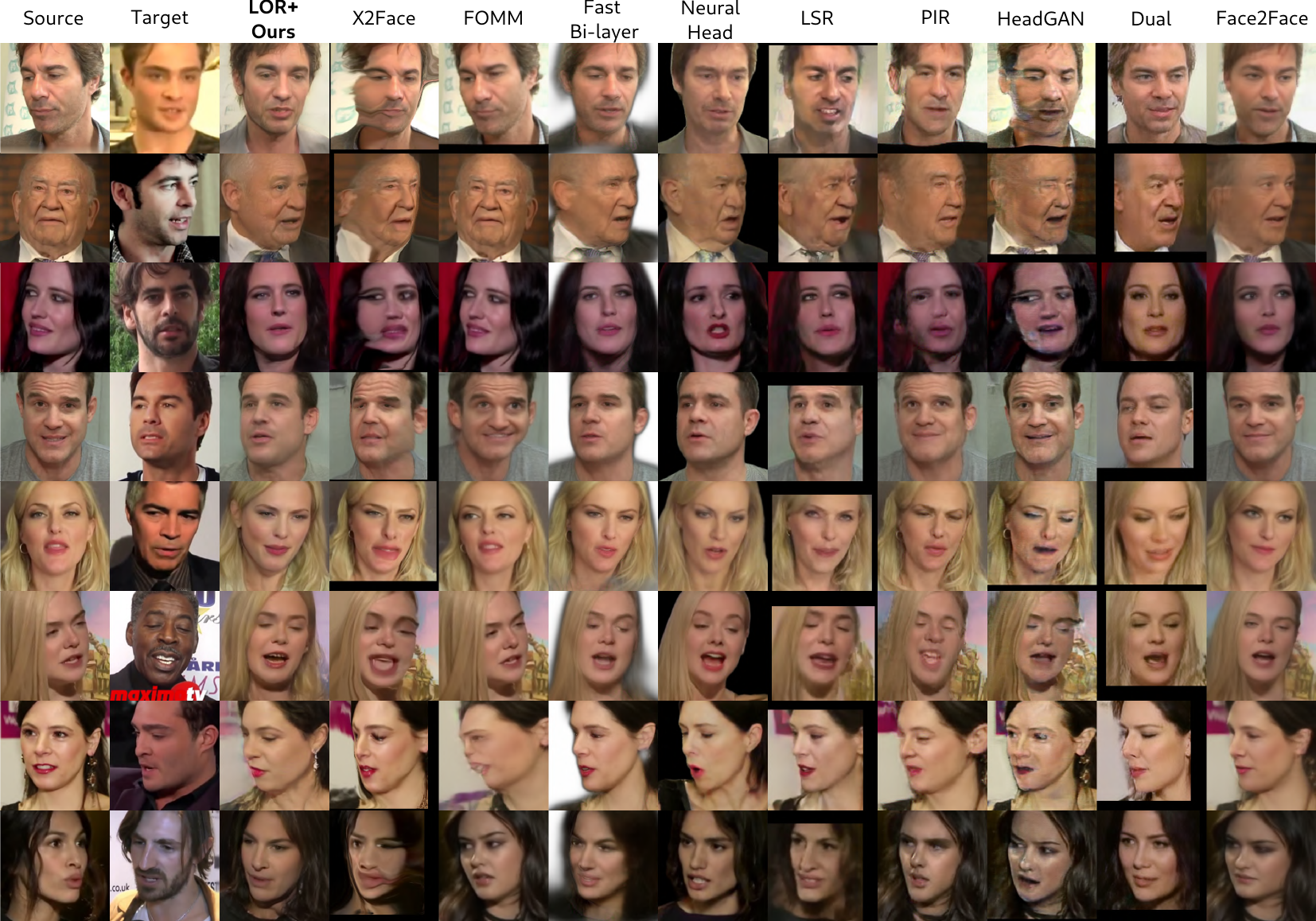}}
\end{center}
  \caption{Qualitative results and comparisons for cross-subject reenactment on VoxCeleb1~\cite{Nagrani17} dataset. The first and second columns show the source and target faces. We compare our method against X2Face~\cite{wiles2018x2face}, FOMM~\cite{siarohin2019first}, Fast Bi-layer~\cite{zakharov2020fast}, Neural-Head~\cite{burkov2020neural}, LSR~\cite{meshry2021learned}, PIR~\cite{ren2021pirenderer}, HeadGAN~\cite{doukas2020headgan}, Dual~\cite{hsu2022dual} and Face2Face~\cite{yang2022face2face}.}\label{fig:cross_vox1_2}
\end{figure*}

\begin{figure*}
\begin{center}
{\includegraphics[width=1.0\linewidth]{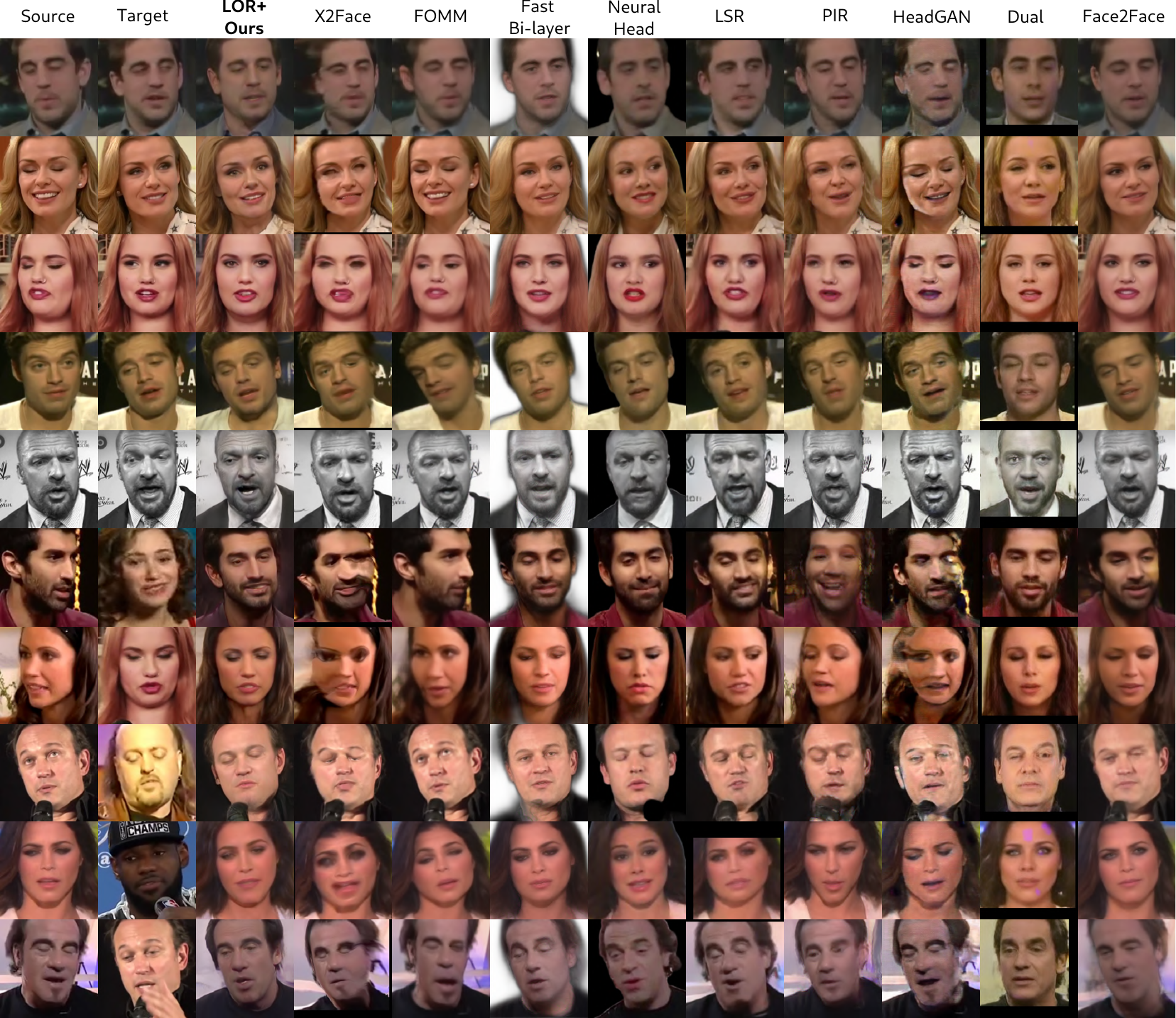}}
\end{center}
  \caption{Qualitative results and comparisons for the tasks of self (first 5 rows) and cross-subject (last 5 rows) reenactment on VoxCeleb2~\cite{Chung18b} dataset. The first and second columns show the source and target faces. We compare our method against X2Face~\cite{wiles2018x2face}, FOMM~\cite{siarohin2019first}, Fast Bi-layer~\cite{zakharov2020fast}, Neural-Head~\cite{burkov2020neural}, LSR~\cite{meshry2021learned}, PIR~\cite{ren2021pirenderer}, HeadGAN~\cite{doukas2020headgan}, Dual~\cite{hsu2022dual} and Face2Face~\cite{yang2022face2face}.}\label{fig:vox2_results}
\end{figure*}

\begin{figure*}[ht]
\begin{subfigure}{.49\textwidth}
  \centering
  % include first image
  \includegraphics[width=.8\linewidth]{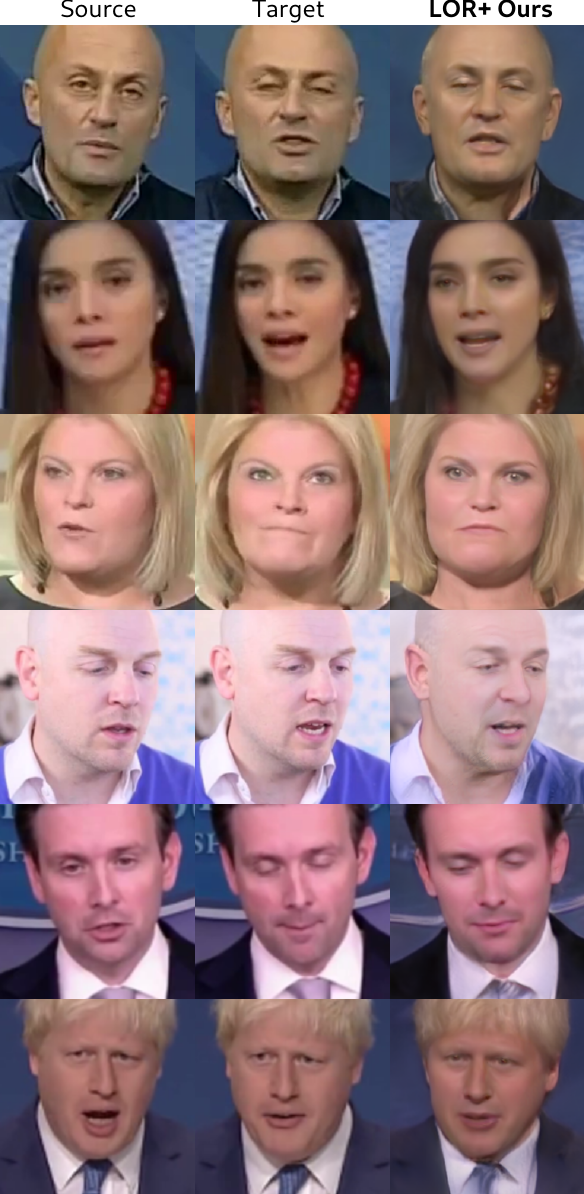}  
  \caption{Self-reenactment.}
  \label{fig:self-sub}
\end{subfigure}
\begin{subfigure}{.49\textwidth}
  \centering
  % include second image
  \includegraphics[width=.8\linewidth]{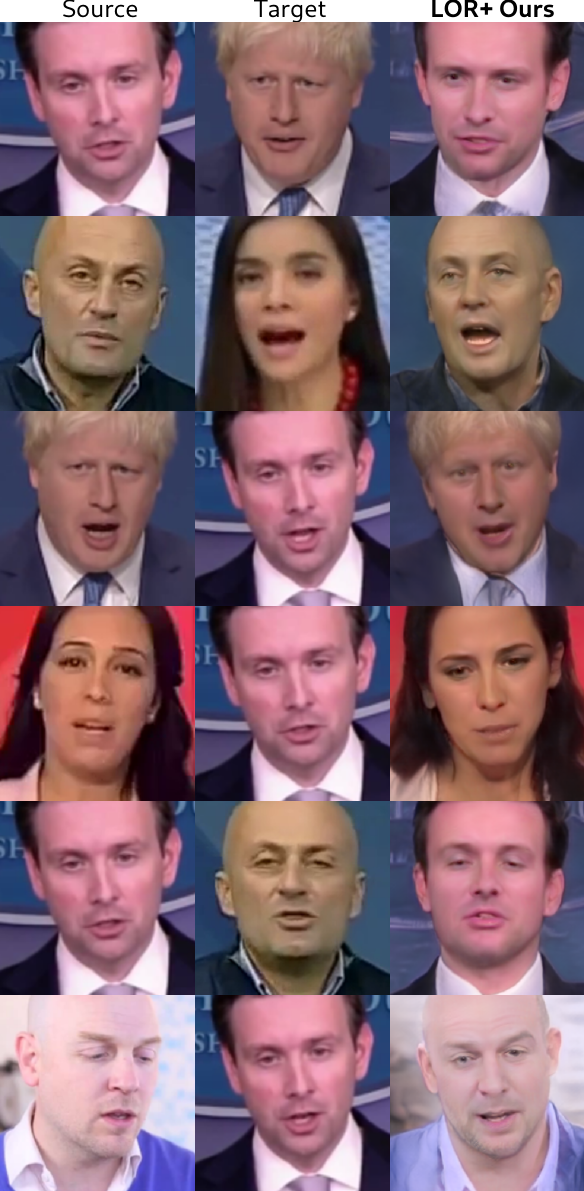}  
  \caption{Cross-subject reenactment.}
  \label{fig:cross-sub}
\end{subfigure}
\caption{Qualitative results of our method for self (a) and cross-subject (b) reenactment on FaceForensics~\cite{roessler2018faceforensics} and 300-VW~\cite{shen2015first} datasets.}
\label{fig:faceforensics}
\end{figure*}

        We provide additional results on self (Fig.~\ref{fig:self_vox1}) and cross-subject (Figs.~\ref{fig:cross_vox1},~\ref{fig:cross_vox1_2}) reenactment on VoxCeleb1~\cite{Nagrani17} dataset and we compare our method with X2Face~\cite{wiles2018x2face}, FOMM~\cite{siarohin2019first}, Fast bi-layer~\cite{zakharov2020fast}, Neural-Head~\cite{burkov2020neural}, LSR~\cite{meshry2021learned}, PIR~\cite{ren2021pirenderer}, HeadGAN~\cite{doukas2020headgan}, Dual~\cite{hsu2022dual} and Face2Face~\cite{yang2022face2face}. Moreover, in Fig.~\ref{fig:vox2_results} we show additional comparisons on VoxCeleb2~\cite{Chung18b} dataset both on self and on cross-subject reenactment. Additionally, we provide a supplementary video with randomly selected videos on self-reenactment and randomly selected pairs on cross-subject reenactment from the test sets of VoxCeleb1 and VoxCeleb2 datasets. Finally, we show that our method is able to generalise well on other facial video datasets. In Fig.~\ref{fig:faceforensics} we provide results on FaceForensics~\cite{roessler2018faceforensics} and 300-VW~\cite{shen2015first} datasets both on self (Fig.~\ref{fig:self-sub}) and on cross-subject (Fig.~\ref{fig:cross-sub}) reenactment.

        \begin{table}[h]
        \caption{Results of a user study that we conduct to evaluate the user preference (Pref. ($\%$)) on the generated images of state-of-the-art methods.}\label{table:user_study}
        % \begin{center}
        \begin{tabular}{|c|c|}
            \hline
            Method & Pref. ($\%$)  \\
            \hline
            X2Face~\cite{wiles2018x2face} & 1.3\\
            FOMM~\cite{siarohin2019first} & 5.0\\
            Fast Bi-layer~\cite{zakharov2020fast} & 9.4\\
            Neural-Head~\cite{burkov2020neural} & \underline{19.2}\\
            LSR~\cite{meshry2021learned} & 10.7\\
            PIR~\cite{ren2021pirenderer}  & 2.3\\
            LOR+ (Ours) & \textbf{52.1}\\ 
            \hline
        \end{tabular}
        % \end{center}
        \end{table}
            
%%%%%%%%%%%%% Comparisons %%%%%%%%%%%%%%%%

\end{appendices}

% \newpage
% \bibliographystyle{plainnat}
\bibliography{ref}% common bib file

\end{document}